\documentclass[11pt]{article}
\usepackage[margin=1in]{geometry}
\usepackage{amsmath,amsfonts,amssymb}
\usepackage{graphicx}
\usepackage{hyperref}
\usepackage{subcaption} 
\usepackage{wrapfig} 
\usepackage{amsmath,amssymb} 
\usepackage{booktabs} 
\usepackage{adjustbox} 
\usepackage{makecell} 
\usepackage{multirow}
\usepackage{caption}
\usepackage{xcolor}
\usepackage[table]{xcolor} 

\newcommand{\instA}{\thanks{Institute of Robotics and Machine Intelligence, Poznan University of Technology}}

\title{Improving Machine Learning-Based Robot Self-Collision Checking with Input Positional Encoding}
\author{Bartłomiej Kulecki\instA ~ \href{https://orcid.org/0000-0002-2820-8212}{ORCID: 0000-0002-2820-8212} \and
Dominik Belter\footnotemark[1] ~ \href{https://orcid.org/0000-0003-3002-9747}{ORCID: 0000-0003-3002-9747}}
\date{}

\begin{document}
\maketitle

\begin{abstract}
This manuscript investigates the integration of positional encoding -- a technique widely used in computer graphics -- into the input vector of a binary classification model for self-collision detection. The results demonstrate the benefits of incorporating positional encoding, which enhances classification accuracy by enabling the model to better capture high-frequency variations, leading to a more detailed and precise representation of complex collision patterns. The manuscript shows that machine learning-based techniques, such as lightweight multilayer perceptrons (MLPs) operating in a low-dimensional feature space, offer a faster alternative for collision checking than traditional methods that rely on geometric approaches, such as triangle-to-triangle intersection tests and Bounding Volume Hierarchies (BVH) for mesh-based models.
\end{abstract}



\bigskip
\noindent\textbf{Disclaimer:} This is a post-peer-review, pre-copyedit version of an article published in:

\noindent B Kulecki, D Belter, Improving Machine Learning-Based Robot Self-Collision Checking with Input Positional Encoding, Foundations of Computing and Decision Sciences, Vol. 50(3), 383-402, 2025

\noindent The final authenticated version is available online at:

\url{https://doi.org/10.2478/fcds-2025-0015}.

\bigskip

\section{Introduction}

\begin{figure}[!t!]
    \centering
    \includegraphics[width=0.9\columnwidth]{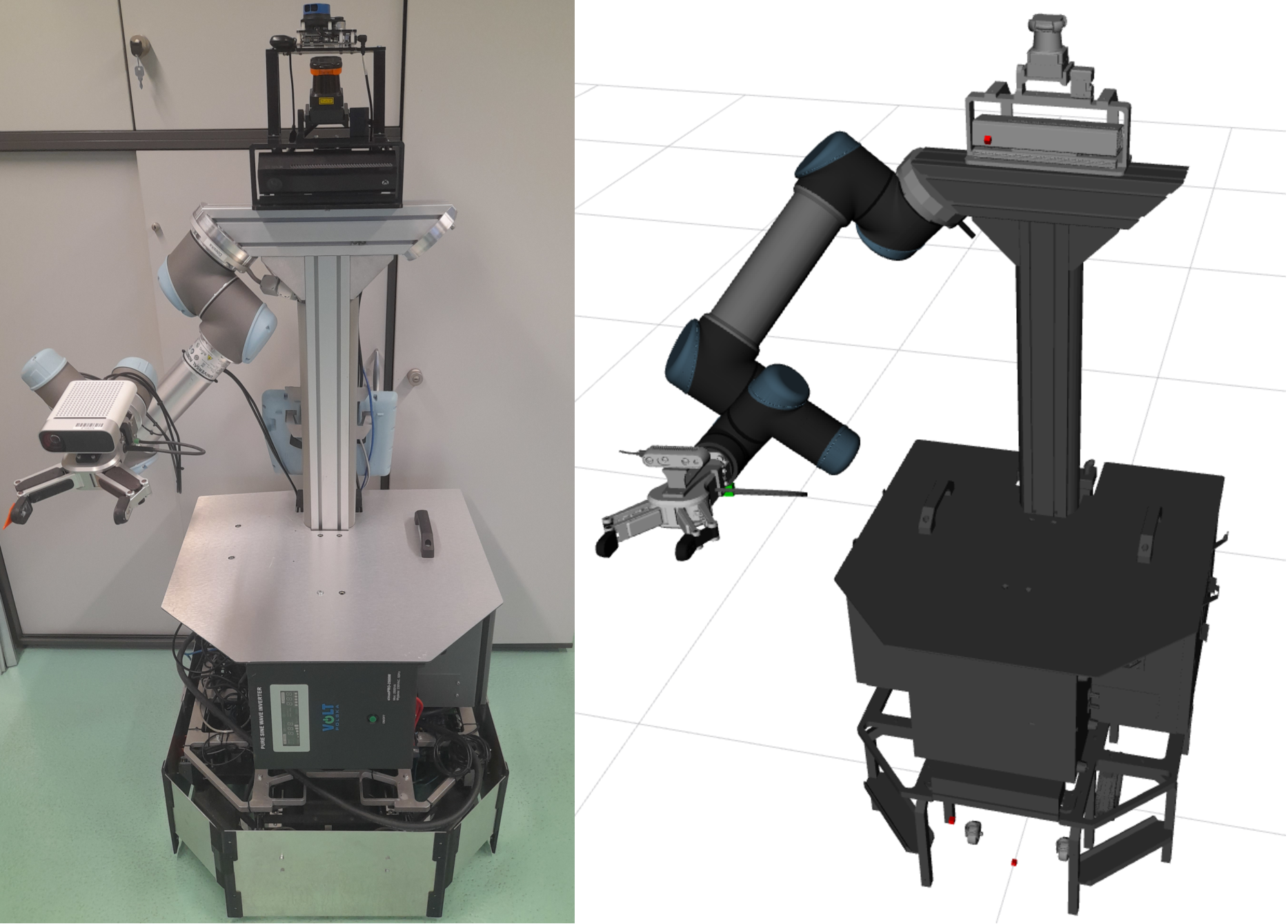}
    \caption{Robot 4.0 and the CAD model of the robot used for implementation of the neural self-collision checking method.}
    \label{fig:robot40}
\end{figure}

Classical collision checking in robotics relies on bounding volumes -- simplified geometric representations such as spheres, cylinders, and boxes -- as well as high-precision mesh models. 
Using simplified 3D models results in fast but less accurate collision detection. Bounding volumes tend to overestimate collision regions by enclosing empty space, which can result in false positives (detecting collisions that do not actually occur) and overly conservative collision checks. As a result, this approach reduces the size of the admissible workspace and limits robot capabilities.
The most accurate collision models use precise mesh models of the robot. 
The Robot Operating System (ROS) uses the Flexible Collision Library (FCL) to detect collisions between meshes~\cite{Pan2012}. FCL employs a two-phase approach: a broad phase for coarse collision detection, followed by a narrow phase for precise triangle-to-triangle intersection tests when necessary. For highly detailed models with a large number of triangles, collision checking can become computationally expensive. To mitigate this, the broad phase leverages a Bounding Volume Hierarchy (BVH)~\cite{Meister2021}, where simplified bounding volumes are used to efficiently approximate potential collisions. However, the two-phase approach introduces variability in collision-checking time, as the computational cost depends on the proximity of objects. If objects are close, the system performs slower but more accurate triangle-level intersection test.
In this manuscript, we focus on methods based on data with constant query time.

Despite the advances of new and more efficient neural network architectures, the multilayer perceptron (MLP)~\cite{Rosenblatt1958perceptron} has regained attention in robotics and computer vision.  
This model has been successfuly applied to various tasks, such as 3D object modeling using signed distance functions~\cite{Park2019deepsdf}, neural scene representation (NeRF) to render images from new camera perspectives (Novel View Synthesis problem)~\cite{Mildenhall2020nerf}, Simultaneous Localization and Mapping (SLAM)~\cite{Rosinol2023}, and controlling robotic manipulators~\cite{Kicki2024}.
The continued relevance of fully connected neural networks lies in their capacity for fast training and their efficiency in solving a wide range of regression and classification tasks in robotics.

Machine learning methods for collision checking~\cite{Krawczyk2023} in robotics typically take a low-dimensional input vector to compute the desired output. This vector can represent different elements, such as the robot's configuration~\cite{Krawczyk2023}, its initial and goal states~\cite{Kicki2024}, or position and viewing direction in 3D space~\cite{Mildenhall2020nerf}. This approach differs from common computer vision tasks, where the input is multidimensional, and the challenge lies in feature extraction. In this study, we aim to enhance the input vector with additional features to improve the neural network's classification accuracy. Specifically, we apply the neural network for self-collision detection in a mobile-manipulating robot~\cite{Krawczyk2023}, as shown in Fig.~\ref{fig:robot40}. Our initial results demonstrate that positional encoding boosts performance when using MLP perceptrons~\cite{Kulecki2024}. By focusing on positional encoding, as introduced in~\cite{Mildenhall2020nerf}, we explore whether this technique, commonly used in computer vision and graphics, can also enhance neural network performance for collision checking in robotics.

In this manuscript, we explore the impact of positional encoding on various machine learning techniques for detecting self-collisions in a manipulating robot. We demonstrate that a multilayer perceptron benefits from an extended feature vector, resulting in improved classification accuracy. Additionally, we compare the effects of positional encoding on MLP performance with its impact on other, non-neural machine learning methods.

\section{Related Work}

\subsection{Positional Encoding}

In this manuscript, we focus on the universal and general method of extending the neural network input vector to improve the accuracy of the binary classification. The goal is to increase the capabilities of neural networks to model small details of the input-output relationship. Rahaman et al. demonstrated that neural networks initially learn lower frequency components, while more complex data manifolds help in the learning of higher frequencies~\cite{Rahaman2018}. NeRF’s positional encoding~\cite{Mildenhall2020nerf}, which augments the input feature vector using a combination of trigonometric functions, enhances the network’s capacity to capture high-frequency features and preserve scene details. A faster alternative to NeRF, Instant NGP~\cite{mueller2022instant}, replaces trigonometric functions with multiresolution hash encoding, aiming to accelerate GPU computations and reduce neural network size. Tancik et al.~\cite{Tancik2020} comprehensively explored Fourier features across various computer graphics tasks, showing their effectiveness in improving MLP performance in low-dimensional regression. In this manuscript, we utilize the approach proposed in~\cite{Mildenhall2020nerf}, because it is general and does not require utilizing hardware acceleration using GPU.

Positional encoding comes from the transformer-like architectures applied in Large Language Models (LLMs)~\cite{Vashwani2017}. Positional encoding in LLMs encodes the position of words in a sentence. First, the pair of sine and cosine functions were used. Later, the techniques utilize the relative position in the sentence and linear bias that is trainable (Relative Bias~\cite{Raffel2020}) or derived from mathematical formula (Align and Bias (AliBi)~\cite{Press2022alibi}). The Rotary Position Embedding (RoPE) method~\cite{Su2024rope} encodes positional information with a rotation matrix. In these methods, the role of the positional encoding is to define the position of the embeddings in the sentence. This article explores the potential application of the positional encoding technique to self-collision detection. In contrast to LLM, the position of the variable related to the robot's configuration does not change. Hence, the role of positional encoding is to encode the position in the configuration space, not the position of the variable in the sequence. For further details on machine learning-based self-collision detection, we refer to~\cite{Krawczyk2023}, which discusses recent methods, including using Gaussian mixtures to infer collisions~\cite{Das2020}.

\subsection{Collision Detection in Motion Planning}

Traditional collision detection methods typically rely on object mesh models, but recent advances have introduced trainable, data-driven approaches. Fastron~\cite{Das2020}, for example, demonstrates that trained models can significantly outperform geometric models in speed, even when methods like Bounding Volume Hierarchy are applied, utilizing a quadratic kernel for faster computations. Similarly, the method proposed in~\cite{Belter2019} employs Gaussian kernels trained with Particle Swarm Optimization~\cite{Kennedy95pso} for self-collision detection in a walking robot. A truncated Gaussian Mixture Model~\cite{Park2020} has been proposed to better define the boundary between collision and collision-free states. In~\cite{Das2020RAL}, replacing the robot's configuration on the input with the distances between characteristic points of the robot arm improves accuracy from 75\% to over 95\% for a 7-degree-of-freedom robot. In~\cite{koptev2022neural}, a neural network was used to estimate distances between the joints of a robotic manipulator and nearby obstacles. This approach yielded an RMSE of 1.05 cm for the predicted distances, achieving a 90\% accuracy in collision detection. This article focuses on using an MLP with the robot's configuration as input, further enhanced with positional encoding to achieve similar levels of accuracy.

Recently, differentiable collision detection methods have also been proposed~\cite{Tracy2023,Montaud2023}. In~\cite{Tracy2023}, the collision detection problem is formulated as a convex optimization task, solving for the minimum uniform scaling applied to each primitive before an intersection occurs. Montaud et al.~\cite{Montaud2023} apply randomized smoothing techniques to estimate the gradients of the collision detection process. Both methods are extremely fast, producing results in microseconds, but they are limited to convex shapes. This limitation can be addressed by approximating complex robotic components with simpler convex shapes~\cite{Valouch2023}, though this approach restricts the range of possible robot movements.

\subsection{Approach and Contribution}
In this work, we investigate the application of positional encoding to the input of an MLP to improve the accuracy of self-collision detection in a manipulating robot. This manuscript is an extended version of our approach previously introduced in the conference papers~\cite{Kulecki2024,Kulecki2024romoco} and [2]. Significant additions include extended experiments on new datasets, a thorough analysis of the number of training samples on the generalization capabilities of the proposed method, and finally, the performance evaluation of the proposed method in the robot motion planning system. The main contributions of this article are as follows: 
\begin{enumerate} 
    \item A general approach for improving the accuracy of binary classification across different MLP architectures, 
    \item Experimental validation presenting the role of positional encoding applied to traditional machine learning techniques, 
    \item Improved accuracy of self-collision detection using an MLP-based method~\cite{Krawczyk2023} and computation efficiency applied in robot motion planning. 
\end{enumerate}

\section{Machine Learning-based Collision Checking}

In this study, we utilize various machine-learning techniques to detect self-collisions in a mobile-manipulating robot. The classification models are trained on a dataset containing collision state labels within the manipulator's configuration space. Each sample includes six joint angle values and a corresponding binary collision label. Our baseline approach involves feeding the robot's configuration directly into neural networks and other machine learning models, as described in~\cite{Krawczyk2023}. In this case, the input vector consists of six values. The positional encoding extends this vector with additional values to enhance training. Previous research has shown that extending the input vector with additional features improves multilayer perceptron performance~\cite{Kulecki2024}. Therefore, we analyze the performance of different neural network architectures and other classifiers when the input data is augmented using positional encoding~\cite{Tancik2020}.

\subsection{Positional Encoding}

Positional encoding~\cite{Mildenhall2020nerf} leverages trigonometric functions to augment the input data, allowing the neural network to capture higher-frequency features. In our method, each joint angle value, denoted as $\boldsymbol{\theta}$ (for 6 DoF robotic arm $\boldsymbol{\theta}=[\Theta_1,...,\Theta_6]$), is encoded using the following function:

\begin{equation}
    \gamma(\boldsymbol{\theta}) = (\boldsymbol{\theta}, \; \sin(2^0\pi \boldsymbol{\theta}), \cos(2^0\pi \boldsymbol{\theta}), \; \dots , \; \sin(2^{L-1}\pi \boldsymbol{\theta}), \cos(2^{L-1}\pi \boldsymbol{\theta}) ),
\label{eq:gamma}
\end{equation}

where $L$ represents the expansion range and defines the length of the input vector. It corresponds to the number of $\sin$ and $\cos$ pairs added for each joint angle. The $\gamma(\boldsymbol{\theta})$ vector is used directly as an input to the neural network. When $L=0$, no positional encoding is applied, and the input vector contains only the joint angles. The total length of the input vector for the machine learning model is:

\begin{equation}
n = 6 \cdot (1+2\cdot L)
\label{eq:length}
\end{equation}

\begin{figure*}[t!]
    \centering
    \resizebox{\textwidth}{!}{%
        \includegraphics[height=3cm]{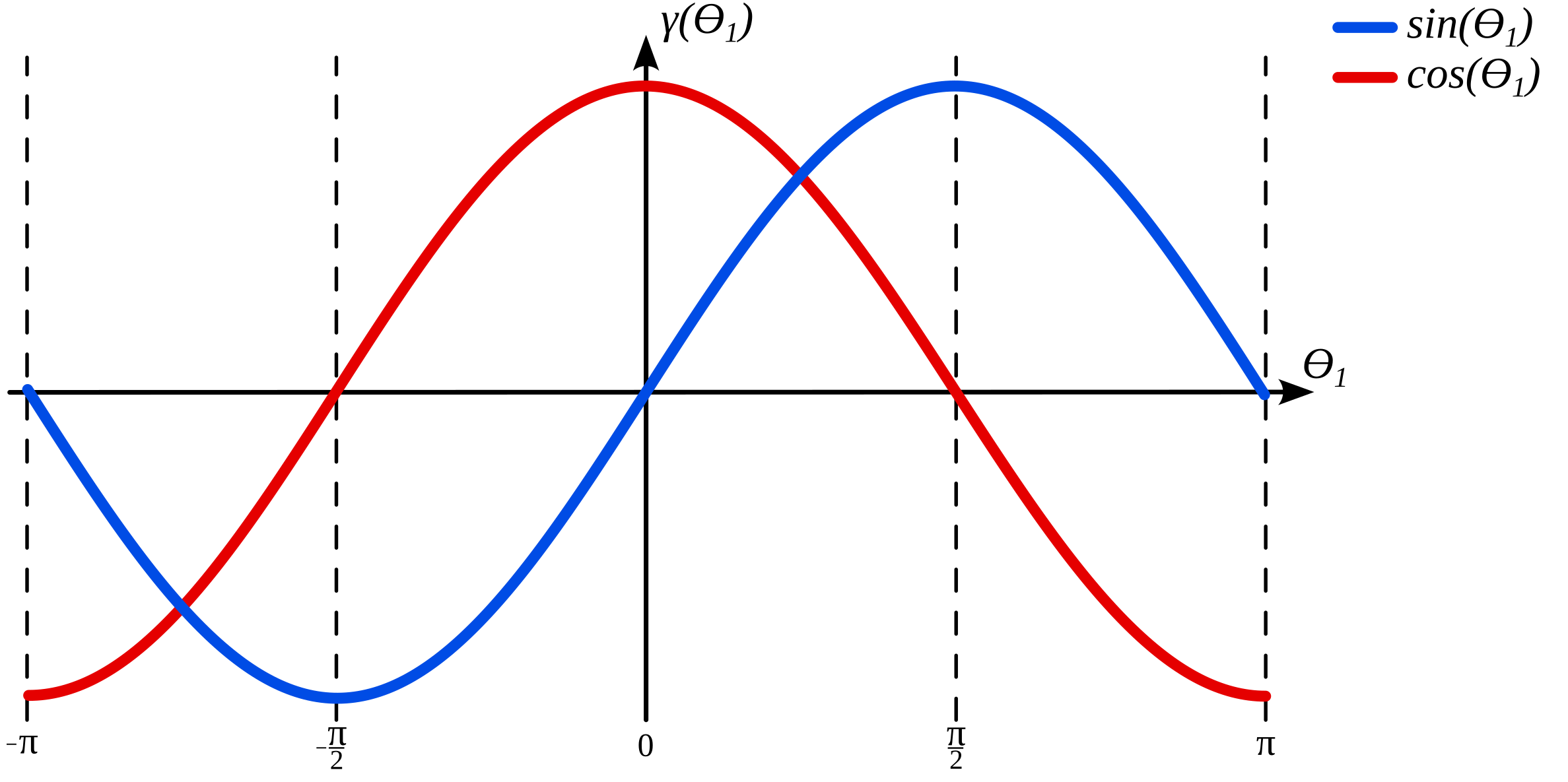}\hspace{1mm}%
        \includegraphics[height=3cm]{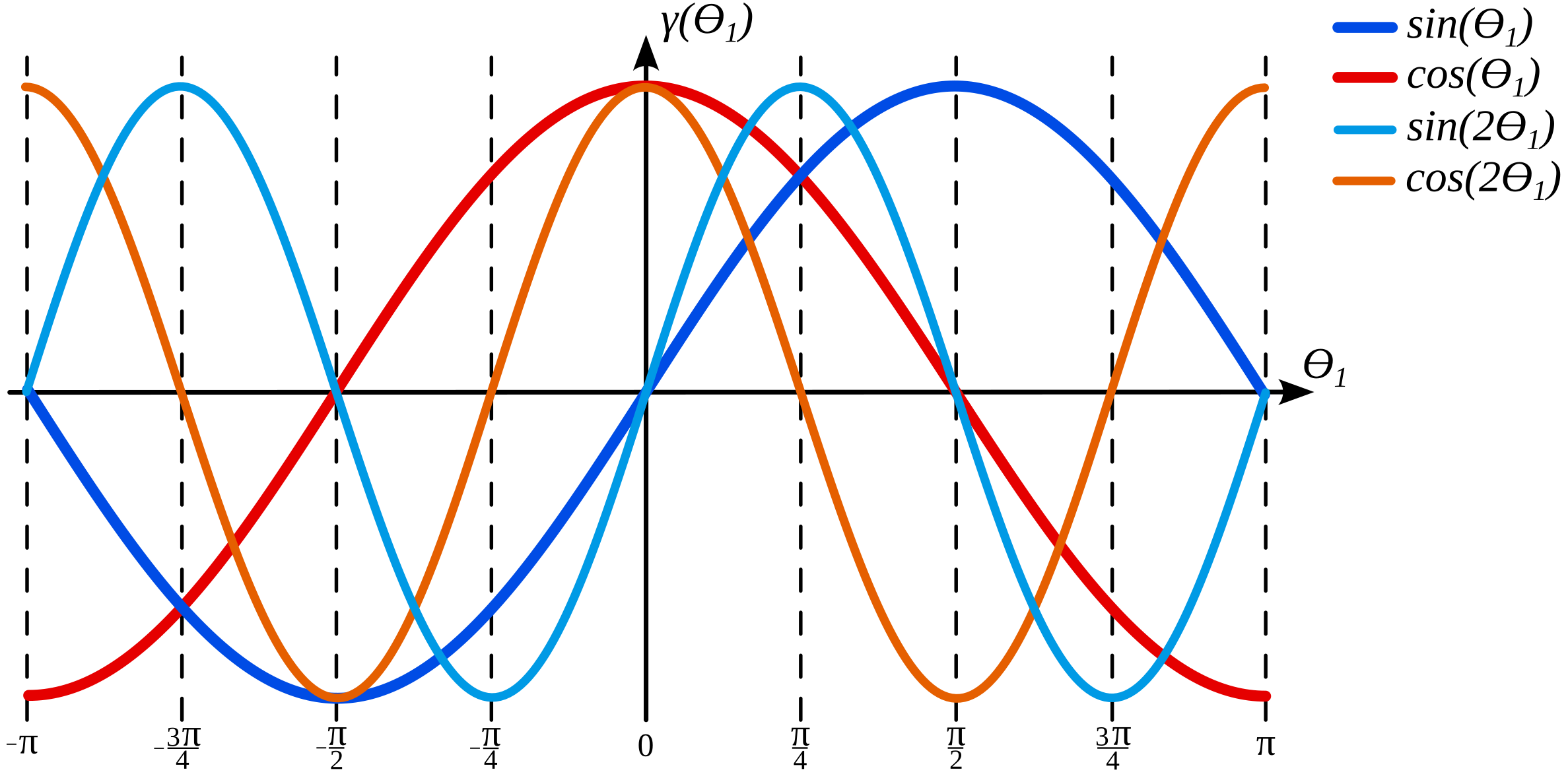}%
    }\put(-368,-9){a} \put(-182,-9){b}\vspace{3mm}
    \resizebox{\textwidth}{!}{%
        \includegraphics[height=3cm]{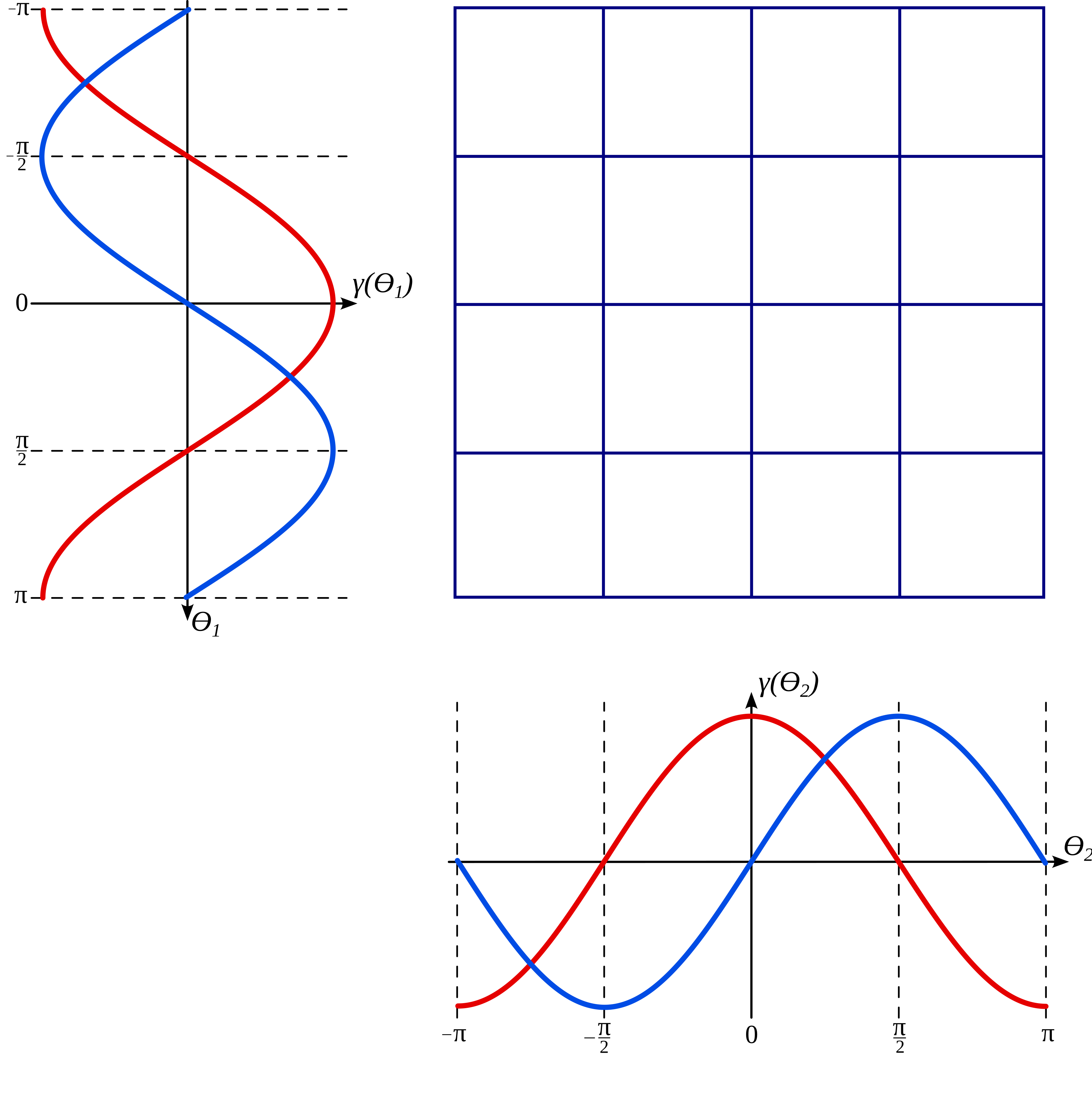}\hspace{1mm}%
        \includegraphics[height=3cm]{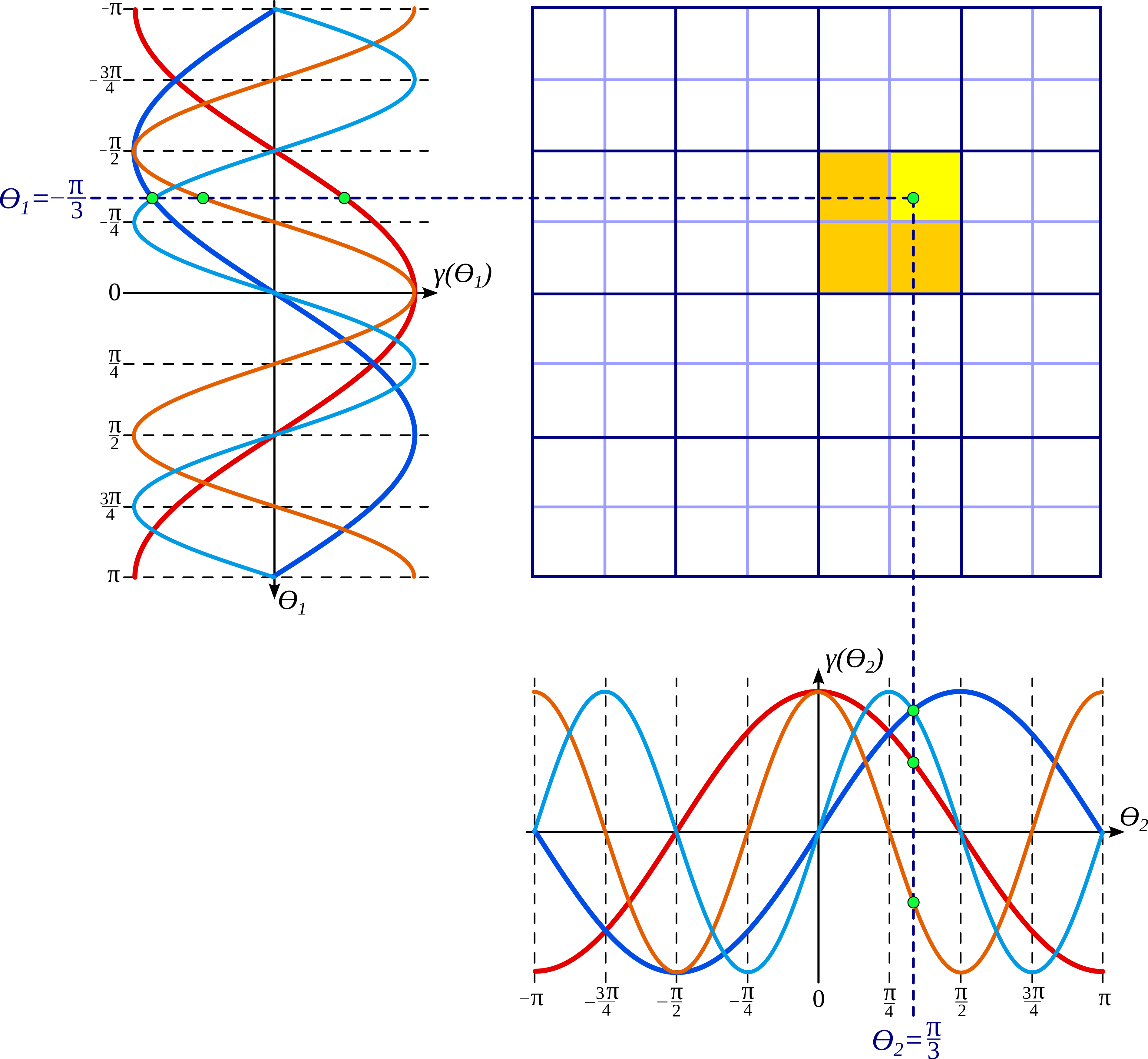}%
    }\put(-368,3){c} \put(-182,3){d}
    \caption{The configuration space divided into ranges with unique encoding values sign combinations: for a single variable, L=1 (a), for a single variable, L=2 (b), for two variables, L=1 (c), and for two variables, L=2, with determining the example input-related region (d).}
    \label{fig:pos-enc-plots}
\end{figure*}

Positional encoding enhances the information included in the input vector. Such a representation makes it easier for the neural network to make precise inferences. The $\sin$ and $\cos$ functions, typically over a $2\pi$ period, produce continuous values that alternate between positive and negative. For a single joint angle $\theta_1$, the pair $\sin(\theta_1)$ and $\cos(\theta_1)$ can be divided into four distinct sign combinations: [+,+], [+,-], [-,-], and [-,+], as illustrated in Fig.~\ref{fig:pos-enc-plots}a. Applying the same to a second joint angle $\theta_2$ results in overlapping sign regions, forming a 2D grid with 4x4 sign combinations, presented in Fig.~\ref{fig:pos-enc-plots}c. Extending this to six joint angles creates a 6D grid with $4^6 = 4096$ unique sign combinations of $\sin$ and $\cos$. By increasing the encoding parameter $L$, additional frequencies are introduced, producing denser encoding grids (Fig.~\ref{fig:pos-enc-plots}b and~\ref{fig:pos-enc-plots}d). This enriched input vector represents joint angles and also identifies the specific cell (or position) within the configuration space where the joint configuration resides, as shown in Fig.~\ref{fig:pos-enc-plots}d. Furthermore, these cells not only inform about the sign but also provide continuous values, supporting precise output computation near the boundaries of these regions.

\subsection{Machine learning models}

We implemented various machine learning models to assess the impact of positional encoding on collision detection. The first model type is a fully connected neural network (MLP), a binary classifier with an input layer matching the size of the input vector and an output layer consisting of a single neuron that typically outputs the probability of class membership. The model's core consists of hidden layers, whose number and size may vary. In addition to the regular MLPs, we utilized the NeRF~\cite{Mildenhall2020nerf} model, as it originally incorporates positional encoding. To evaluate its effectiveness, we tested both the original NeRF and our modified version. The neural network architectures we employed in this study (illustrated in Fig.~\ref{fig:architectures}) are as follows:

\begin{figure}[!t!]
    \centering
    \includegraphics[width=\columnwidth]{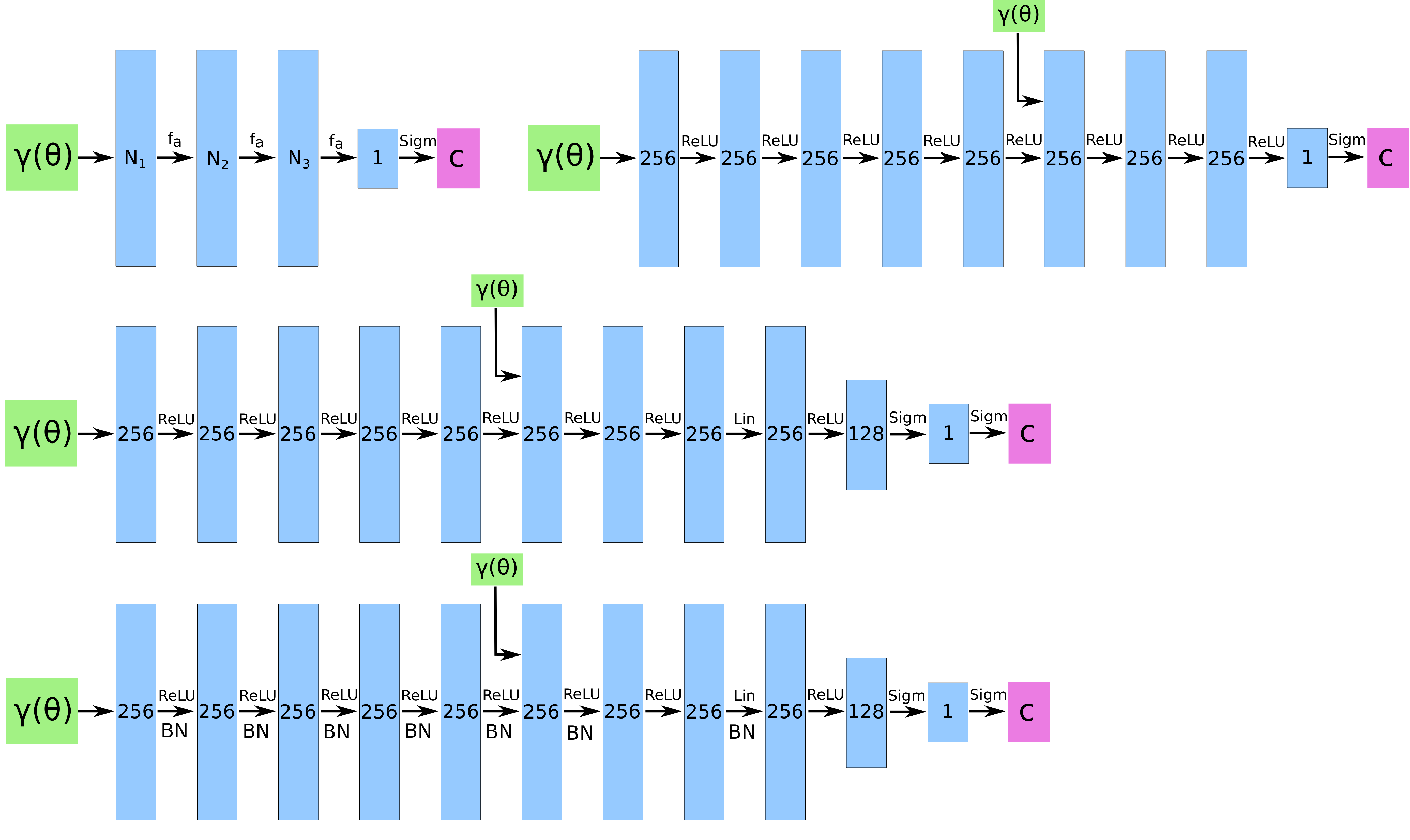}
    \put(-368,201){a} \put(-232,201){b}
    \put(-368,130){c} \put(-368,55){d}
    \caption{MLP architectures used in this research: MLP resulting from auto-sklearn optimization (a), original NeRF architecture (b), our implementation of NeRF architecture (c), and our implementation of NeRF with batch normalization (d).}
    \label{fig:architectures}
\end{figure}

\begin{itemize} 
\item \textbf{MLP} – the MLP classifier from the scikit-learn library (Fig.~\ref{fig:architectures}a). The auto-sklearn framework was applied with various MLP input configurations to optimize hyperparameters such as the number of hidden layers, sizes of the hidden layers ($N_1$, $N_2$, $N_3$), activation function $f_a$, and learning rate.

\item \textbf{NeRF~\cite{Mildenhall2020nerf}} – the original NeRF implementation (Fig.~\ref{fig:architectures}b), featuring 8 hidden layers, each with 256 neurons. ReLU (Rectified Linear Unit) activation is applied after every hidden layer, while the final output layer uses a sigmoid activation function $f_\sigma$. A unique aspect of this architecture is the skip connection, which concatenates the input vector to the 6th hidden layer.

\item \textbf{NeRF\textsubscript{MLP}} – our custom implementation of the NeRF architecture based on the original paper~\cite{Mildenhall2020nerf} (Fig.~\ref{fig:architectures}c). This network consists of 10 hidden layers: the first 9 layers with 256 neurons and the last hidden layer with 128 neurons. We employ a skip connection to the 6th layer and ReLU activations. It differs from the original implementation in using a linear activation for the 8th layer and a sigmoid activation for the 10th smaller layer.

\item \textbf{NeRF\textsubscript{MLP} + BN} – another NeRF-based implementation (Fig.~\ref{fig:architectures}d), similar to NeRF\textsubscript{MLP} but with the addition of batch normalization (BN) layers. BN is applied before each ReLU-activated layer to stabilize and accelerate training. 
\end{itemize}

The second group of models we employed consists of classical machine learning algorithms: Linear Discriminant Analysis (LDA), Gaussian Naive Bayes (GNB), Random Forest (RF), K-Nearest Neighbors (KNN), Support Vector Machine (SVM), and Gradient Boosting (GB). As with the MLP models, the parameters for each were optimized using the auto-sklearn framework across various input vector configurations.

\begin{figure}[!t!]
    \centering
    \includegraphics[width=0.99\columnwidth]{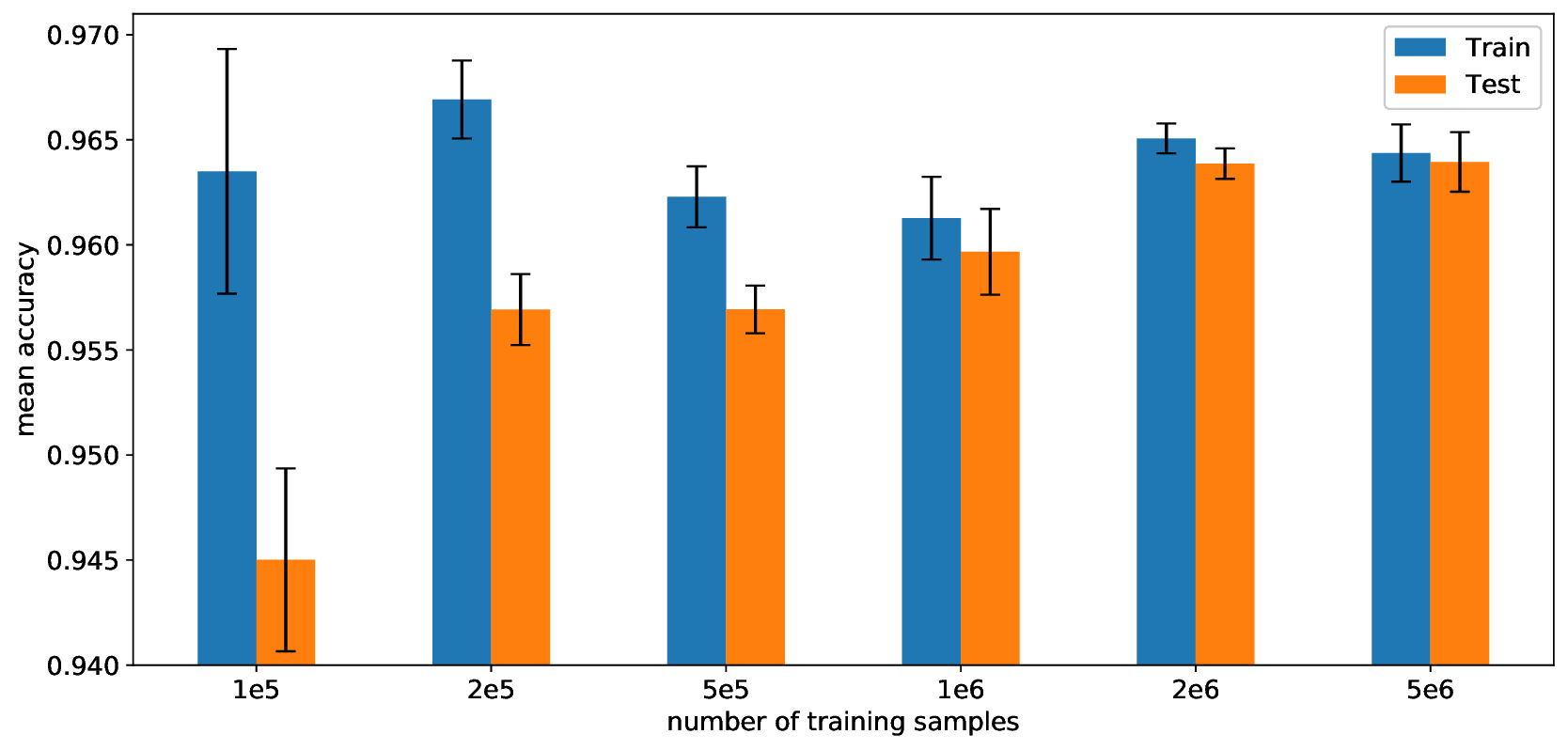}
    \caption{Influence of the number of training samples on the difference between training and testing accuracy. The values represent mean accuracy based on 5 training and testing trials using an MLP classifier consisting of 3 hidden layers with 176 neurons each.}
    \label{fig:acc_vs_datano}
\end{figure}

\subsection{Dataset and training}

The dataset for this study was generated by randomly sampling the manipulator's configuration space using a uniform distribution. Each sample includes $n=6$ input values, uniformly drawn from the range $[-\pi,\pi]$, and a collision state, determined using the Flexible Collision Library (FCL)~\cite{Pan2012}. The dataset was split into training, test, and validation sets in a 70:20:10 ratio. All sets are balanced -- collision and non-collision states are represented in a 1:1 ratio. We checked the difference between the accuracy obtained during testing and training the neural network to determine the number of samples required to train the model. The results are presented in Fig.~\ref{fig:acc_vs_datano}. For 100k training samples, the difference between the training and test accuracy equals 1.41\%. This suggests that the neural network is overfitted. The difference value decreases when the number of training samples increases and reaches 0.05\% for 5 million training samples. 
It is important to consider that training time increases with the size of the dataset. Therefore, a dataset containing 1 million training samples was generated for this study.

We trained the model using the Adam optimizer and binary cross-entropy loss function, running for 200 epochs. Each training and testing procedure was repeated 10 times to calculate the mean metric value and standard deviation. Accuracy, defined as the sum of true positive and true negative classifications divided by the total number of test samples, was used as the evaluation metric. 
We use a non-weighted sum of true positives (TP) and true negatives (TN) as our evaluation metric, following standard practice in binary classification problems. This choice ensures consistency with previous studies and allows for a direct comparison with existing methods.
However, in real-world robotic motion planning, false negatives (FN), where a collision occurs but is not detected, are significantly more critical than false positives (FP), which only result in avoiding some feasible configurations.

\begin{figure*}[t]
    \centering
    \includegraphics[width=0.333\textwidth]{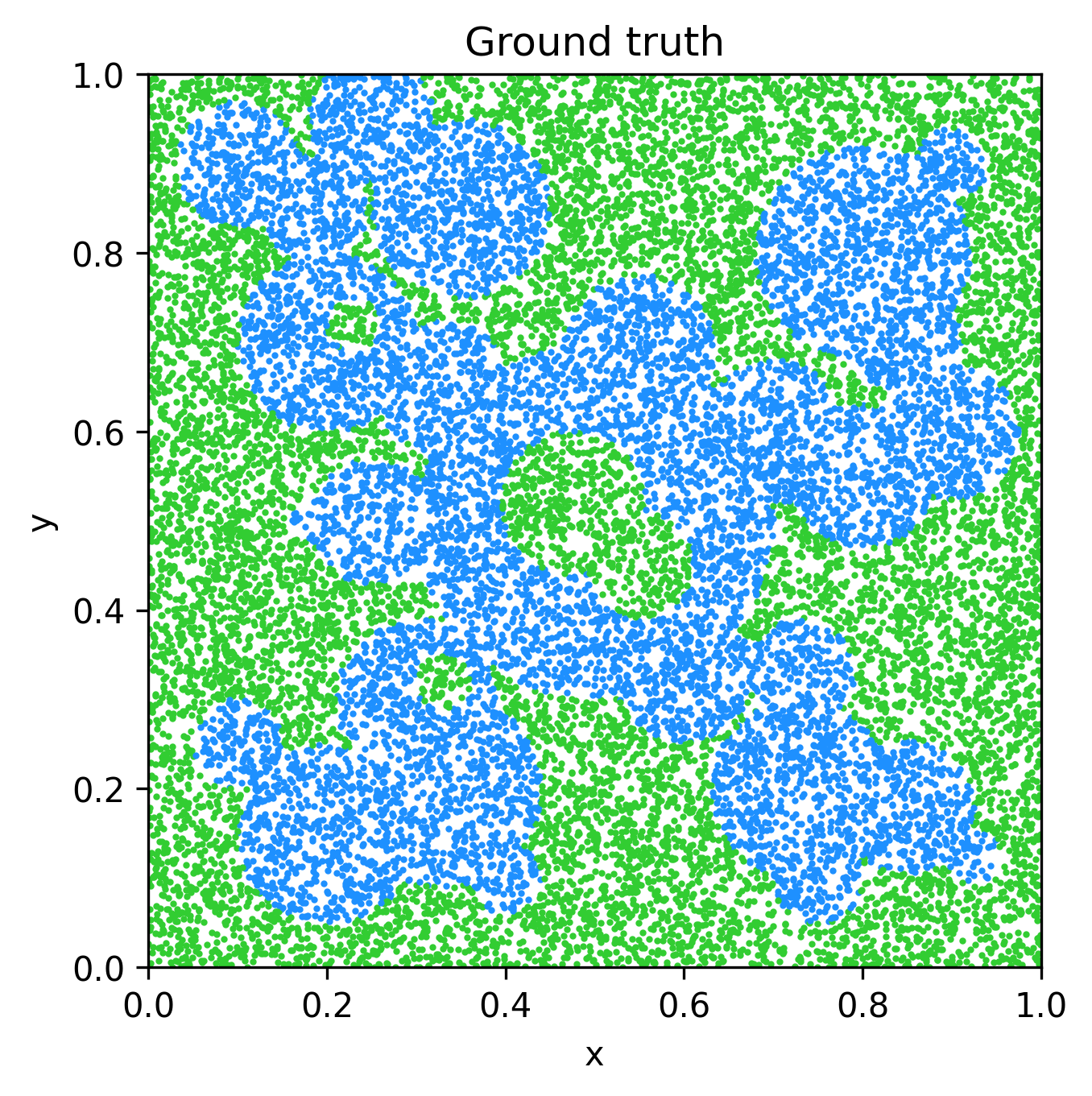}%
    \includegraphics[width=0.333\textwidth]{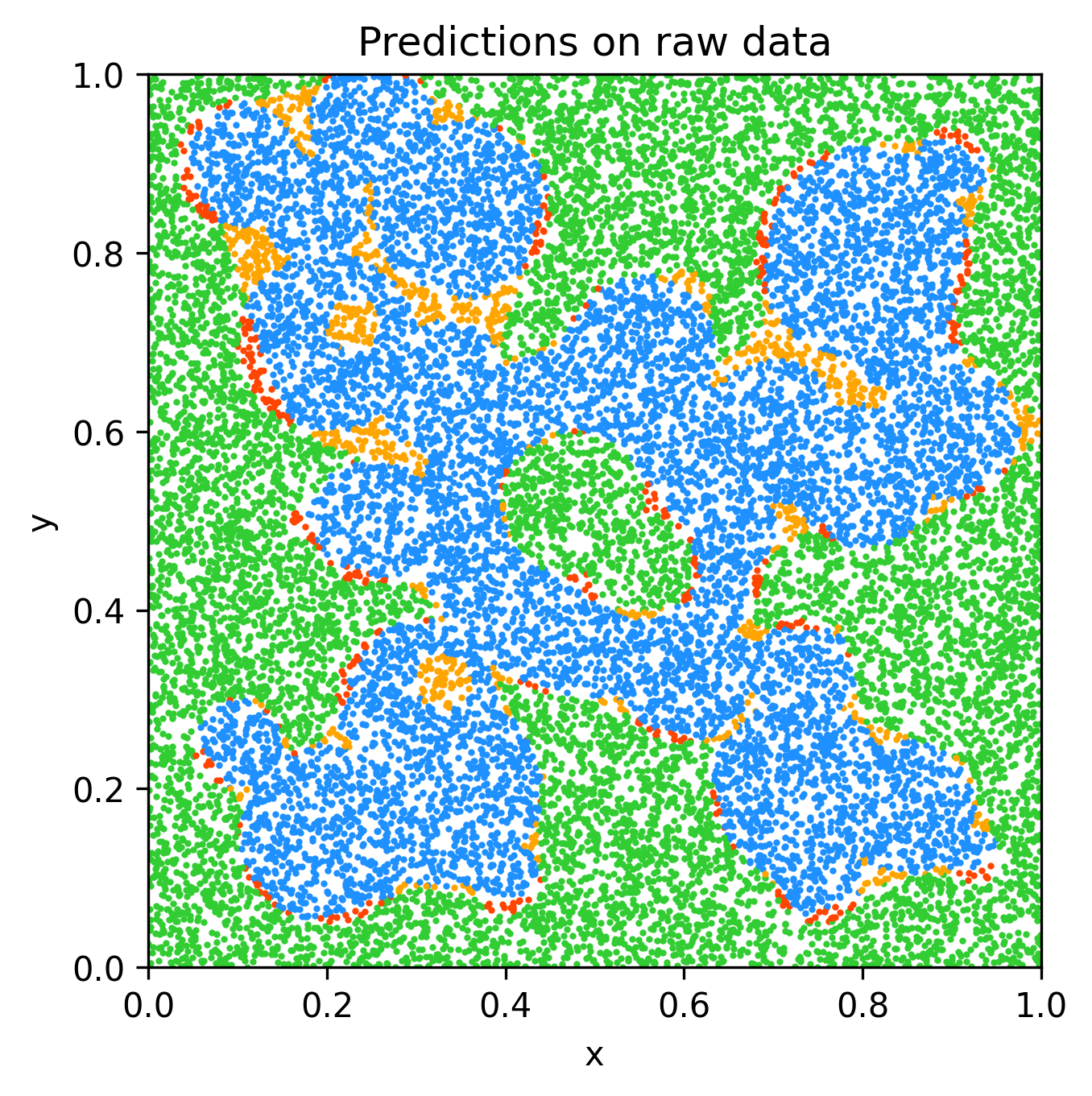}%
    \includegraphics[width=0.333\textwidth]{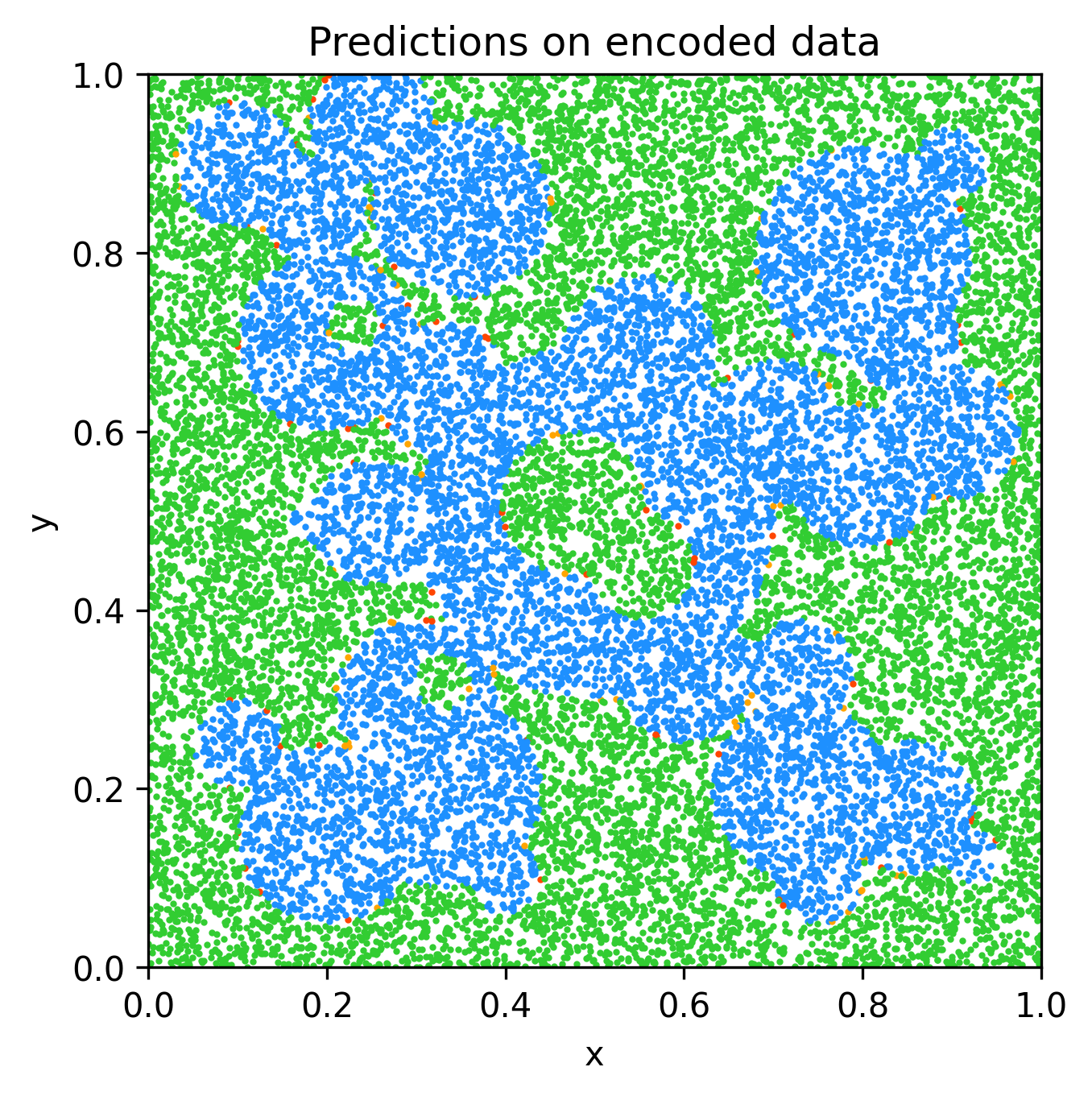}
    \caption{Results of classification in 2D space: ground truth test data, predictions based on raw $x,y$ coordinates, and predictions based on the encoded input vector. Colors: blue -- TP, green -- TN, orange -- FP, red -- FN.}
    \label{fig:circles}
\end{figure*}

\section{Results}

\subsection{Preliminary experiments in 2D space}

The initial experiments focus on verifying whether the proposed input encoding method boosts binary classification accuracy in a low-dimensional space. For this purpose, we utilized artificial data consisting of x and y values ranging from 0.0 to 1.0 and a collision label (0 or 1). We generated 100.000 samples and picked the MLP architecture with two hidden layers of size 32 neurons. The first model was trained using raw data (2 input values), and the second using data encoded with the function $\gamma$ and parameter L = 5 (22 input values). As a result, we observe a significant difference in mean test accuracy: 0.8899 vs 0.9919. Fig.~\ref{fig:circles} illustrates the ground truth and predicted labels in the 2D space. It shows that a neural network trained with the raw data cannot preserve small details and reproduce complicated class boundaries. Training with encoded data enables the simple MLP model to learn more complex functions.

\subsection{Architectures optimization}

To develop efficient models for detecting the robot's self-collisions, optimizing the hyperparameters was a crucial step. Table~\ref{tab:architectures_mlp} outlines the number of neurons in the hidden layers and the activation functions for the MLPs, which were optimized for different input configurations. The optimization process began with the shortest input feature vector without positional encoding ($L=0$), proceeded through a single $\sin, \cos$ pair ($L=1$), and extended to $L=16$. For the optimization process, we utilized a smaller dataset with 200k samples. The auto-sklearn framework consistently returned architectures with three hidden layers, each containing the same number of neurons. Optimization for each model took 6 hours, and accuracy was used as the selection criterion. The same steps were performed for conventional machine learning models to obtain the best parameters.

\begin{table}[t]
\caption{Number of neurons in the following layers ($N_1$, $N_2$, $N_3$) of the MLP obtained after optimization for various L parameter values.}
\label{tab:architectures_mlp}
\vspace{-0.1cm}
\begin{center}
\setlength\tabcolsep{3.0pt}
\begin{tabular}{c|ccccccc}
architecture name & MLP0 & MLP1 & MLP3 & MLP6 & MLP9 & MLP12 & MLP16 \\ \hline
$L$    &  0  & 1  & 3 & 6 & 9 & 12 & 16  \\ \hline
$N_1$  &  106 & 176 & 51 & 38 & 25 & 160 & 111   \\ 
$N_2$  &  106 & 176 & 51 & 38 & 25 & 160 & 111 \\ 
$N_3$  &  106 & 176 & 51 & 38 & 25 & 160 & 111 \\
$f_a$  &  tanh & ReLU & ReLU & ReLU & tanh & tanh & tanh 
\vspace{-3mm}
\end{tabular}
\end{center}
\end{table}

\subsection{Basic MLP architectures}

Tab.~\ref{tab:comparison-mlp} shows the results of training the model to detect the robot's collision state, comparing inputs based solely on configuration (not encoded) to those enhanced with positional encoding. We tested several values of $L$, but only the best mean accuracy for encoded input is presented in the table. In all cases, the incorporation of positional encoding improved classification accuracy. While the mean accuracy gain is near 1\%, it's important to note that these architectures have already performed well even without positional encoding.

\begin{figure*}[t!]
    \centering
    \includegraphics[width=0.333\textwidth]{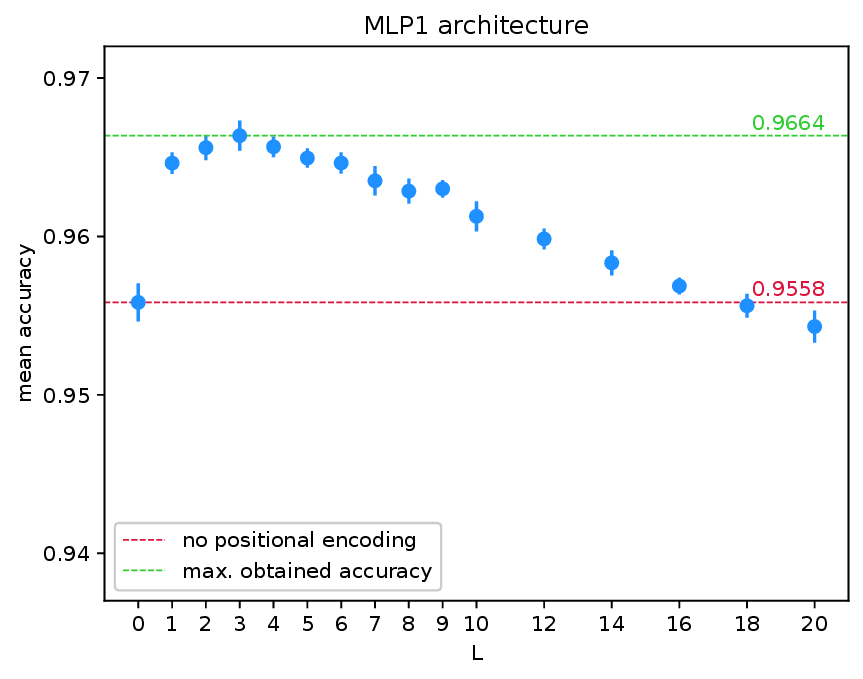}%
    \includegraphics[width=0.333\textwidth]{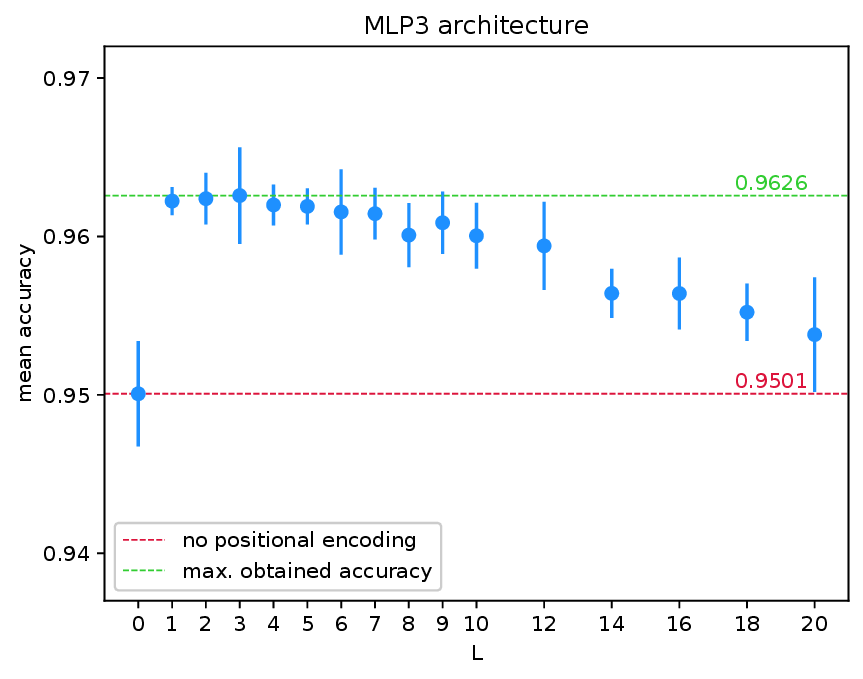}%
    \includegraphics[width=0.333\textwidth]{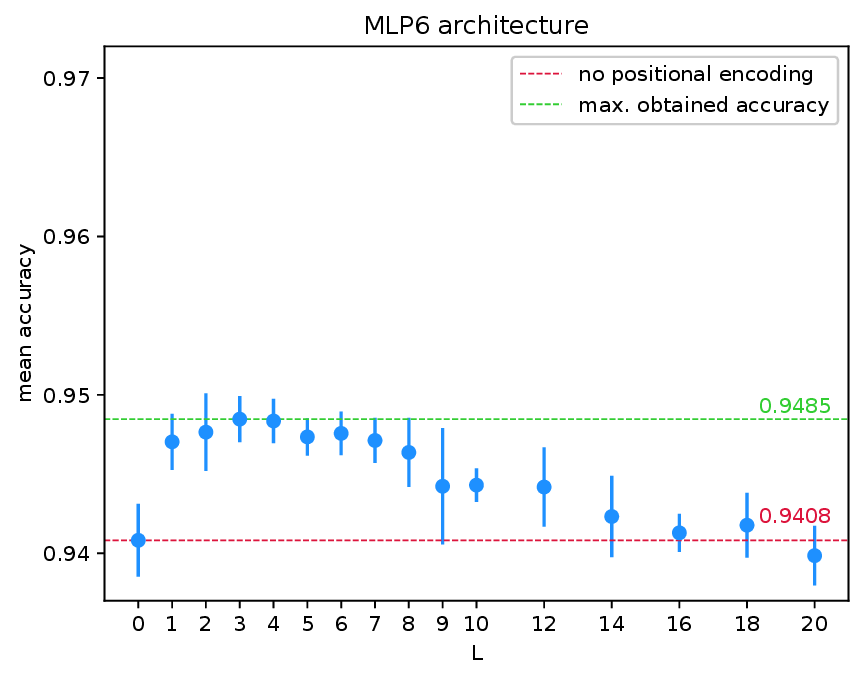}
    \includegraphics[width=0.333\textwidth]{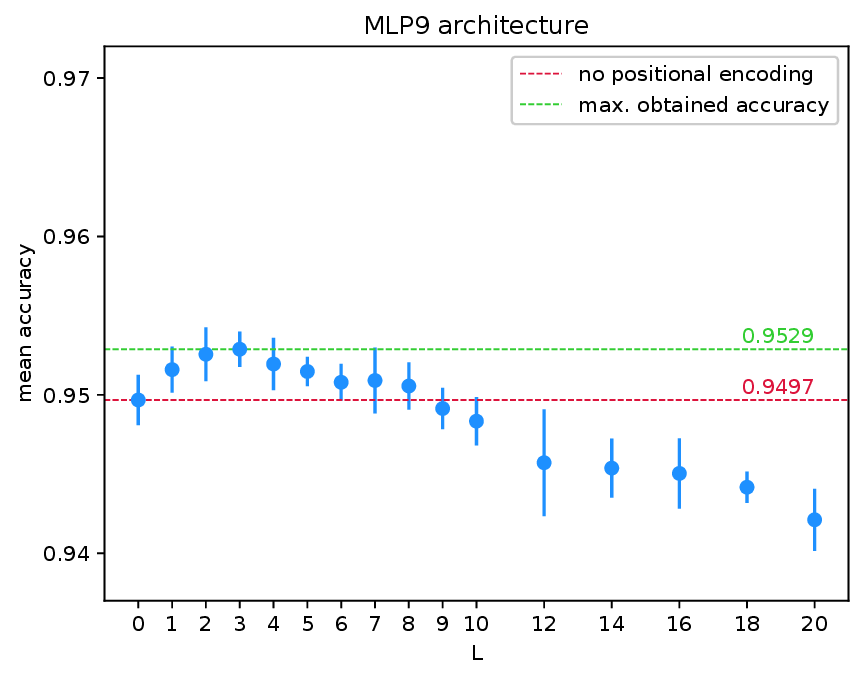}%
    \includegraphics[width=0.333\textwidth]{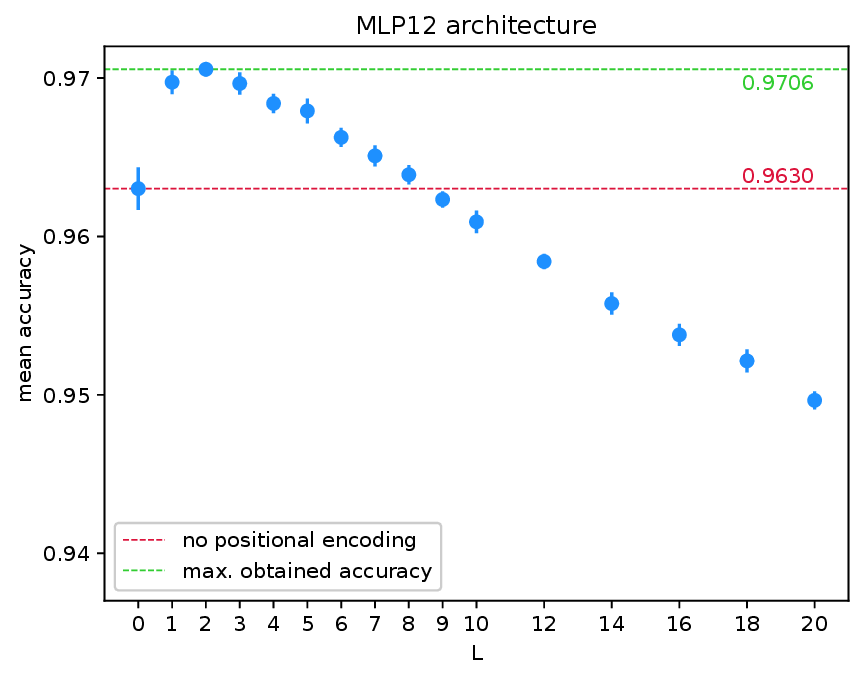}%
    \includegraphics[width=0.333\textwidth]{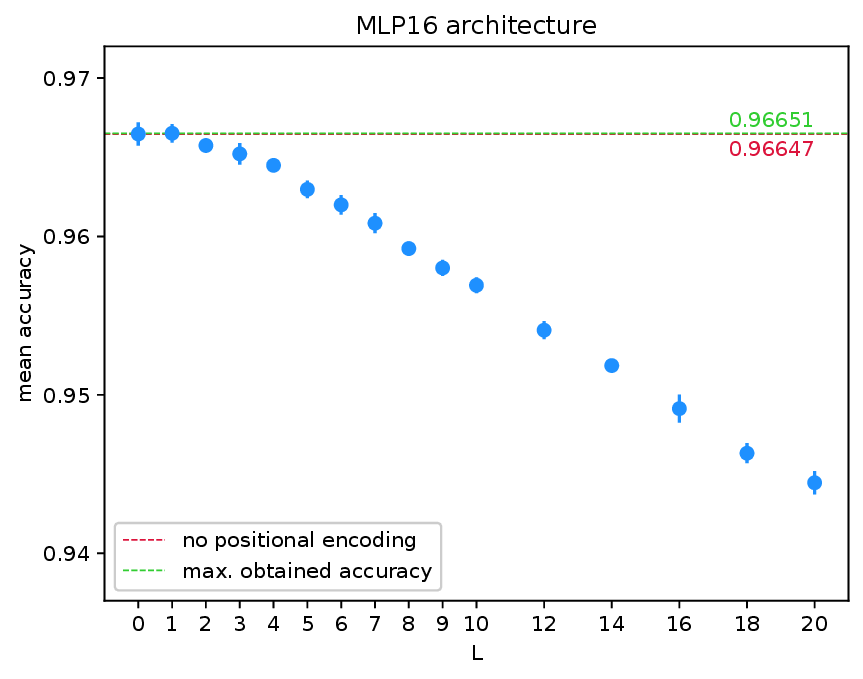}
    \caption{Mean classification accuracy for various positional encoding $L$ parameter values obtained for selected MLP architectures.}
    \label{fig:accuracy-mlp}
\end{figure*}

All optimized MLP architectures were trained on datasets, both without positional encoding ($L=0$) and with augmented data for $L$ values from 1 to 20. This resulted in input vector sizes ranging from $n=6$ for $L=0$ to $n=246$ for $L=20$.
Figure~\ref{fig:accuracy-mlp} illustrates the results across all $L$ values. The red line represents the accuracy for $L=0$ (without positional encoding), and the green line marks the best accuracy achieved with positional encoding. The highest accuracy is typically observed for $L$ values between 1 and 3, after which a further increase in input vector length leads to a near-linear decline in accuracy. The architecture optimized for $L=12$ (MLP12) achieved the best accuracy, reaching 0.9706. However, as the input vector length grows, the benefit of positional encoding is reduced. Results for MLP0 and MLP16 show that input augmentation does not enhance prediction performance significantly for those neural network architectures.

\begin{table*}[t]
\caption{Best of mean classification accuracy obtained with and without positional encoding for selected MLP architectures.}
\label{tab:comparison-mlp}
\vspace{-0.1cm}
\begin{center}
\setlength\tabcolsep{2.0pt}
\begin{tabular}{c|ccccccc}
architecture   & MLP0  & MLP1  & MLP3  & MLP6   & MLP9 & MLP12 & MLP16 \\ \hline
without pos. enc.    & 0.9667 & 0.9558& 0.9501& 0.9408&  0.9497& 0.9630& {\bf 0.9665}\\ 
with pos. enc.       & {\bf 0.9668}& {\bf 0.9664}& {\bf 0.9626}& {\bf 0.9485}& {\bf 0.9529}& {\bf 0.9706}& {\bf 0.9665}\\ 
improvement [\%] & 0.006& 1.101&  1.317& 0.813& 0.337& 0.782& 0.005\\
best L               & 1& 3&  3& 3& 3& 2& 1
\vspace{-3mm}\end{tabular}
\end{center}
\end{table*}

\subsection{NeRF-based architectures}
In the following experiments, all NeRF-like models were trained using TensorFlow with identical training parameters. For each model, the average accuracy from 10 training and testing trials was calculated, and the results are shown in Tab.~\ref{tab:comparison-nerf}. The mean improvement from adding positional encoding to NeRF-like models is consistent with the enhancements seen in MLPs. However, applying positional encoding to the original NeRF model~\cite{Mildenhall2020nerf} did not yield significant improvements.

\begin{figure*}[t!]
    \centering
    \includegraphics[width=0.333\textwidth]{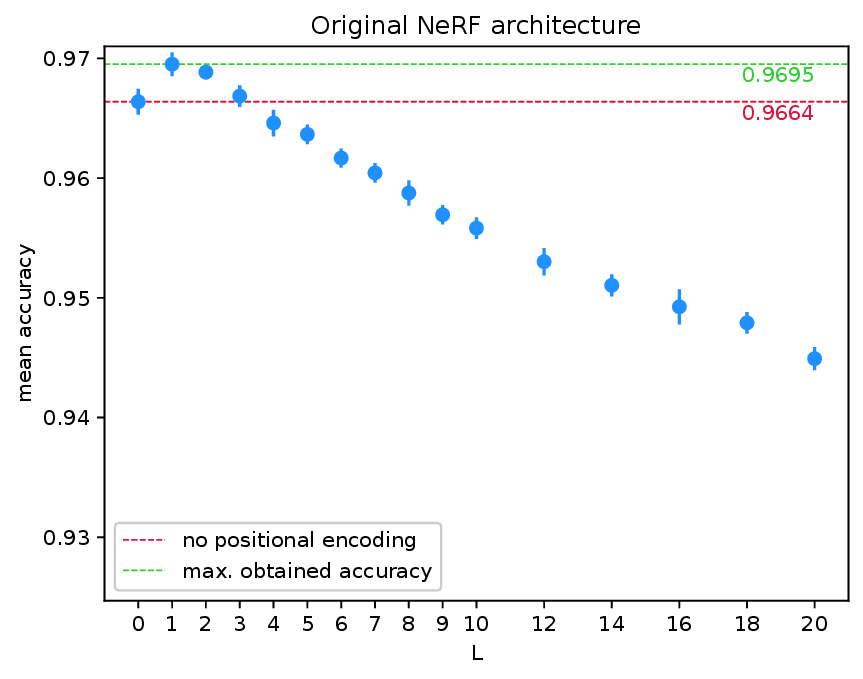}%
    \includegraphics[width=0.333\textwidth]{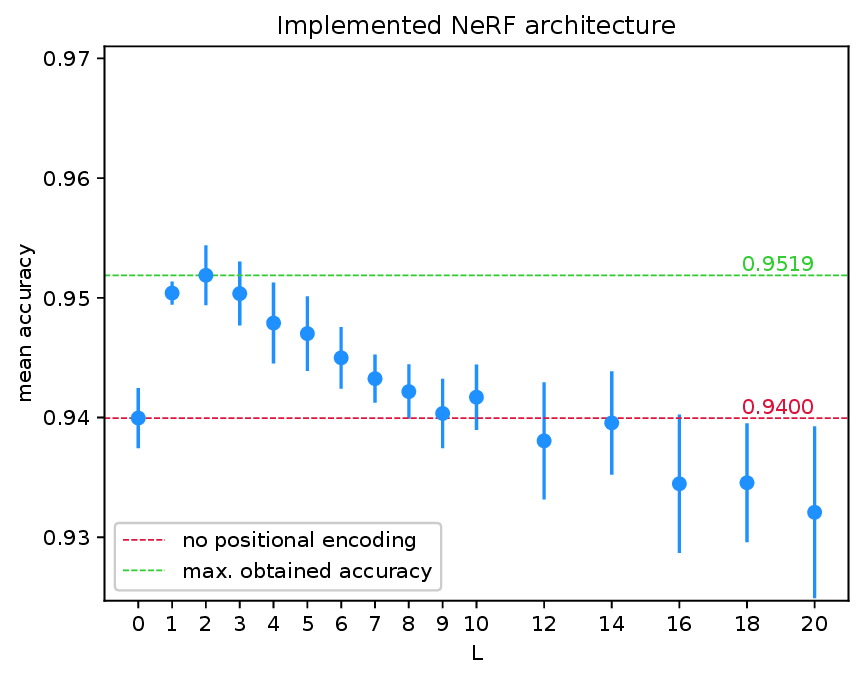}%
    \includegraphics[width=0.333\textwidth]{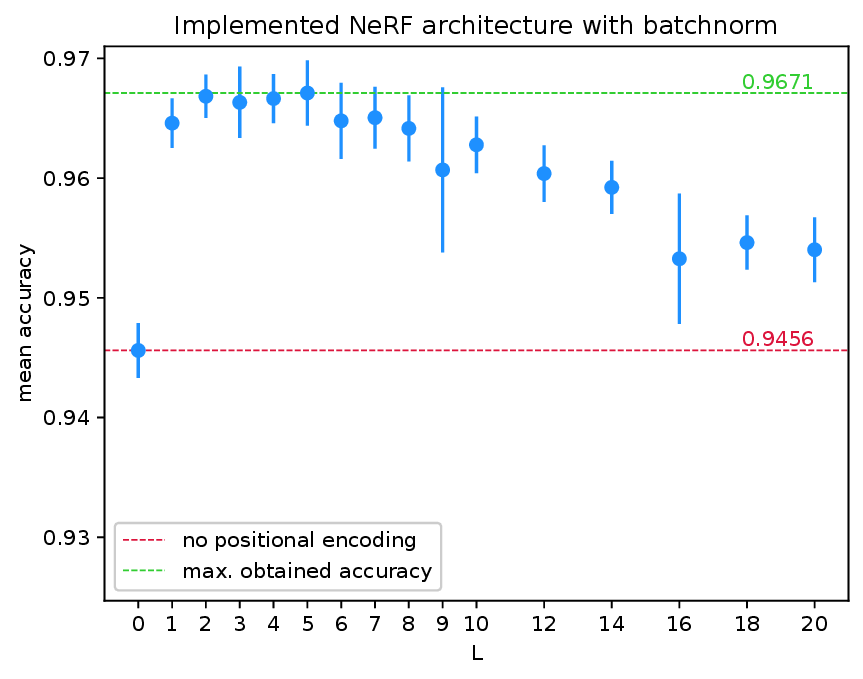}
    \caption{Mean classification accuracy for various positional encoding $L$ parameter values obtained for selected NeRF-like architectures.}
    \label{fig:accuracy-nerf}
\end{figure*}

The ${\rm NeRF}_{\rm MLP}$ model with batch normalization (BN) demonstrated the largest improvement after incorporating input encoding (over 2.27\%). However, the highest accuracy among neural networks is obtained for the original NeRF architecture. The full range of results across different input vector lengths is depicted in Fig.~\ref{fig:accuracy-nerf}. The optimal $L$ value for NeRF models equals 1. It increases to 5 for the ${\rm NeRF}_{\rm MLP}$ model with BN. Results presented in Fig.~\ref{fig:accuracy-nerf} show that accuracy drops when the $L$ value is too high. This indicates that simpler neural networks struggle to extract meaningful information when the input vector becomes overly large. For this problem, it is recommended to keep the $L$ value relatively small (under 5).

\begin{table*}[t]
\caption{Best of mean classification accuracy obtained with and without positional encoding for original NeRF~\cite{Mildenhall2020nerf} and selected architectures of NeRF-like MLPs.}
\label{tab:comparison-nerf}
\vspace{-0.1cm}
\begin{center}
\setlength\tabcolsep{2.5pt}
\begin{tabular}{c|ccc}
architecture  & NeRF~\cite{Mildenhall2020nerf}& ${\rm NeRF}_{\rm MLP}$ & ${\rm NeRF}_{\rm MLP}$ + BN  \\ \hline
without pos. enc.    & 0.9664&  0.9400& 0.9456 \\ 
with pos. enc.       & {\bf 0.9695}& {\bf 0.9519}& {\bf 0.9671}\\ 
improvement [\%] & 0.324& 1.268       & 2.274 \\
best L               & 1           & 2           & 5 
\vspace{-3mm}
\end{tabular}
\end{center}
\end{table*}

\subsection{Conventional machine learning methods}

The final set of experiments aimed to assess the effect of positional encoding on other classical machine learning methods. Again, we used auto-sklearn to optimize the classifiers' parameters for various input vector lengths. The results for models optimized with $L=3$ are shown in Table~\ref{tab:comparison-other}. Positional encoding has a positive impact on nearly all classifiers except Random Forest. The accuracy improvement is significant for lower-performing models such as the LDA and SVM, though their overall accuracy remains low. The best-performing model is Gradient Boosting, achieving an accuracy of 0.9766, outperforming MLPs. However, positional encoding has a small effect on the performance in this case.

\begin{figure*}[t]
    \centering
    \includegraphics[width=0.333\textwidth]{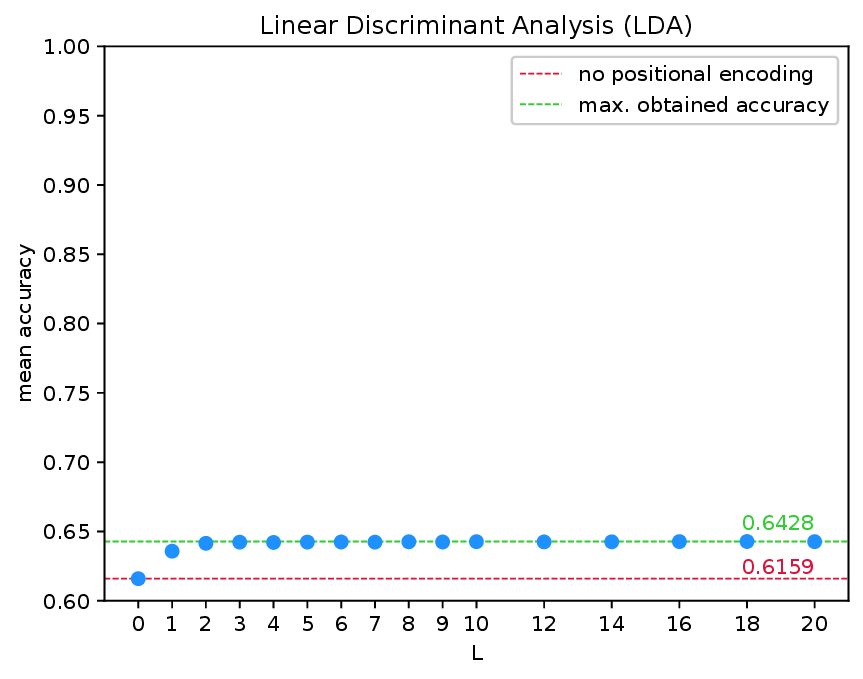}%
    \includegraphics[width=0.333\textwidth]{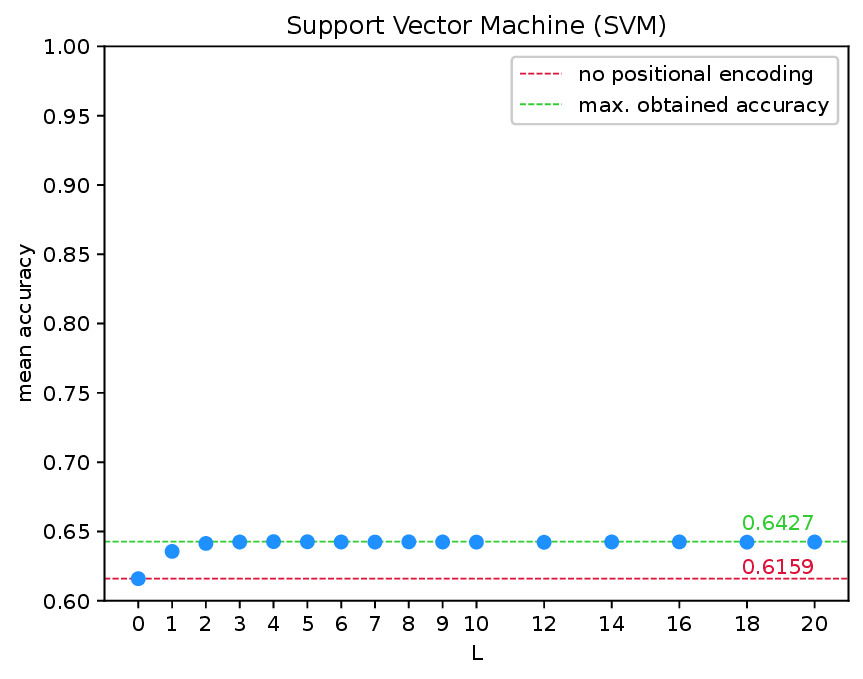}%
    \includegraphics[width=0.333\textwidth]{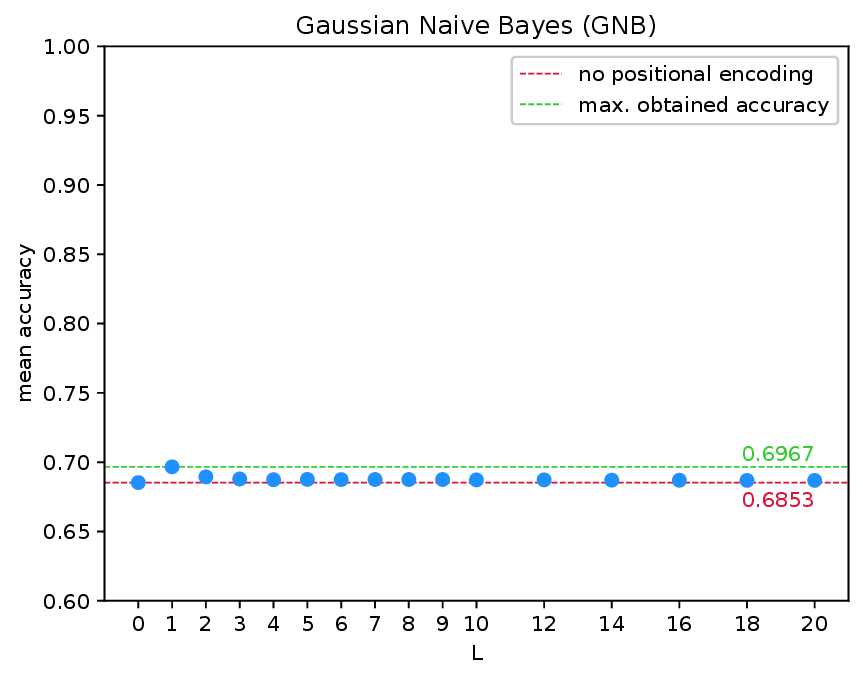}
    \includegraphics[width=0.333\textwidth]{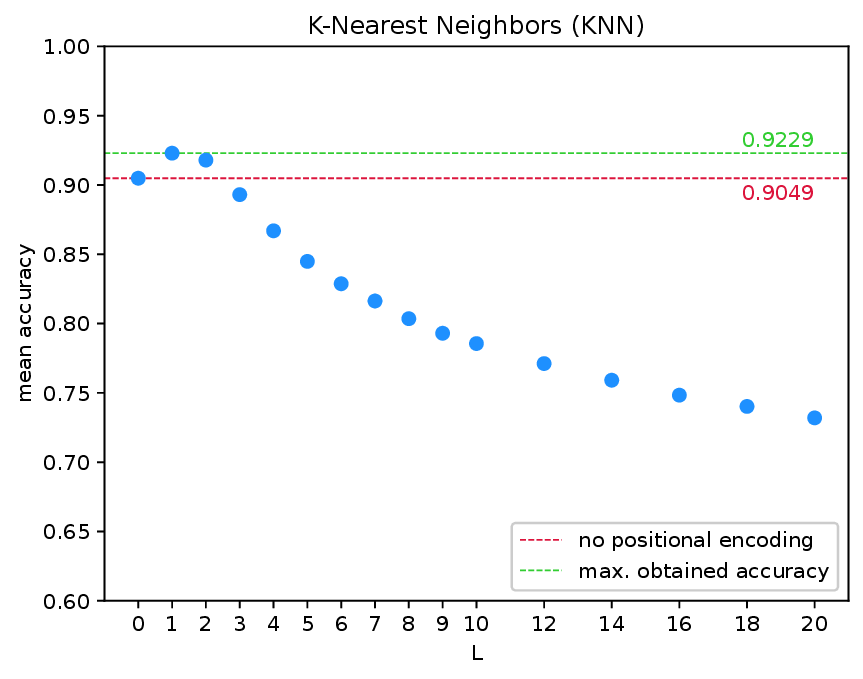}%
    \includegraphics[width=0.333\textwidth]{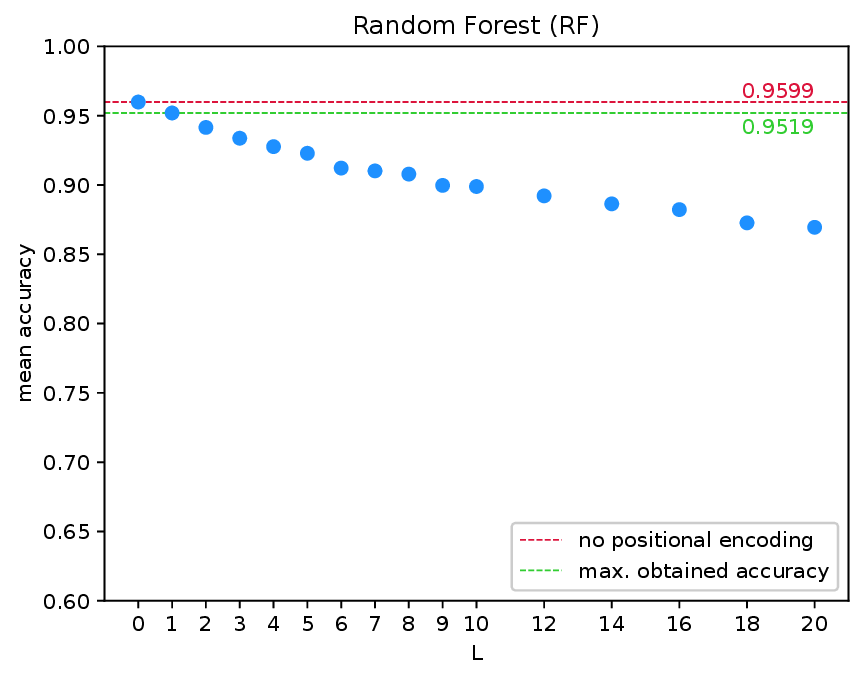}%
    \includegraphics[width=0.333\textwidth]{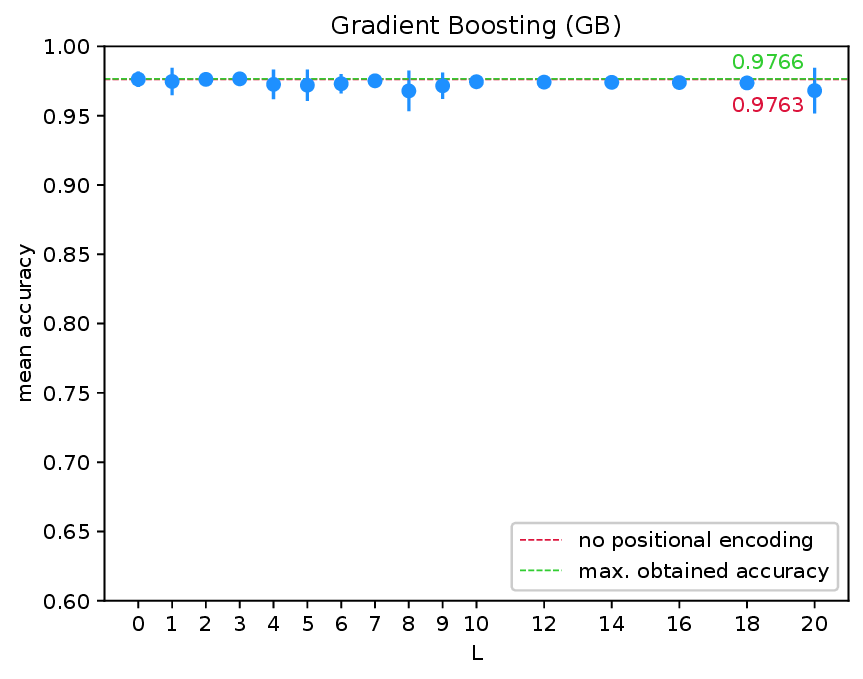}
    \caption{Mean classification accuracy for various positional encoding $L$ parameter values obtained for selected ML classifiers optimized for L=3.}
    \label{fig:accuracy-other}
\end{figure*}

Figure~\ref{fig:accuracy-other} illustrates the relationship between mean accuracy and the degree of input augmentation. The trends for classical models differ somewhat from those observed for neural networks. For models like LDA, SVM, and GNB, enabling positional encoding improves accuracy, but further increasing input length $L$ has little effect. In contrast, for k-NN, accuracy improves at lower $L$ values but drops as input length grows, with accuracy falling below baseline levels beyond $L=3$. For the RF the positional encoding has no positive effect and for the GB method, the improvement is statistically insignificant.

\begin{table}[t]
\caption{Best of mean classification accuracy obtained with and without positional encoding for selected ML classifiers optimized for L=3.}
\label{tab:comparison-other}
\vspace{-0.1cm}
\begin{center}
\setlength\tabcolsep{2.0pt}
\begin{tabular}{c|cccccc}
classifier           & LDA         & SVM         & GNB         & KNN         & RF          & GB  \\ \hline
without pos. enc.    & 0.6159& 0.6159& 0.6853& 0.9049& {\bf 0.9599}& 0.9763\\ 
with pos. enc.       & {\bf 0.6428}& {\bf 0.6427}& {\bf 0.6967}& {\bf 0.9229}& 0.9519& {\bf 0.9766}\\ 
improvement [\%] & 4.354& 4.343& 1.657& 1.997& -0.824& 0.029\\
best L               & 18& 4& 1           & 1           & 1& 3 
\vspace{-3mm}\end{tabular}
\end{center}
\end{table}

\subsection{Qualitative results}

\begin{figure}[t]
    \centering
    \includegraphics[width=1.0\linewidth]{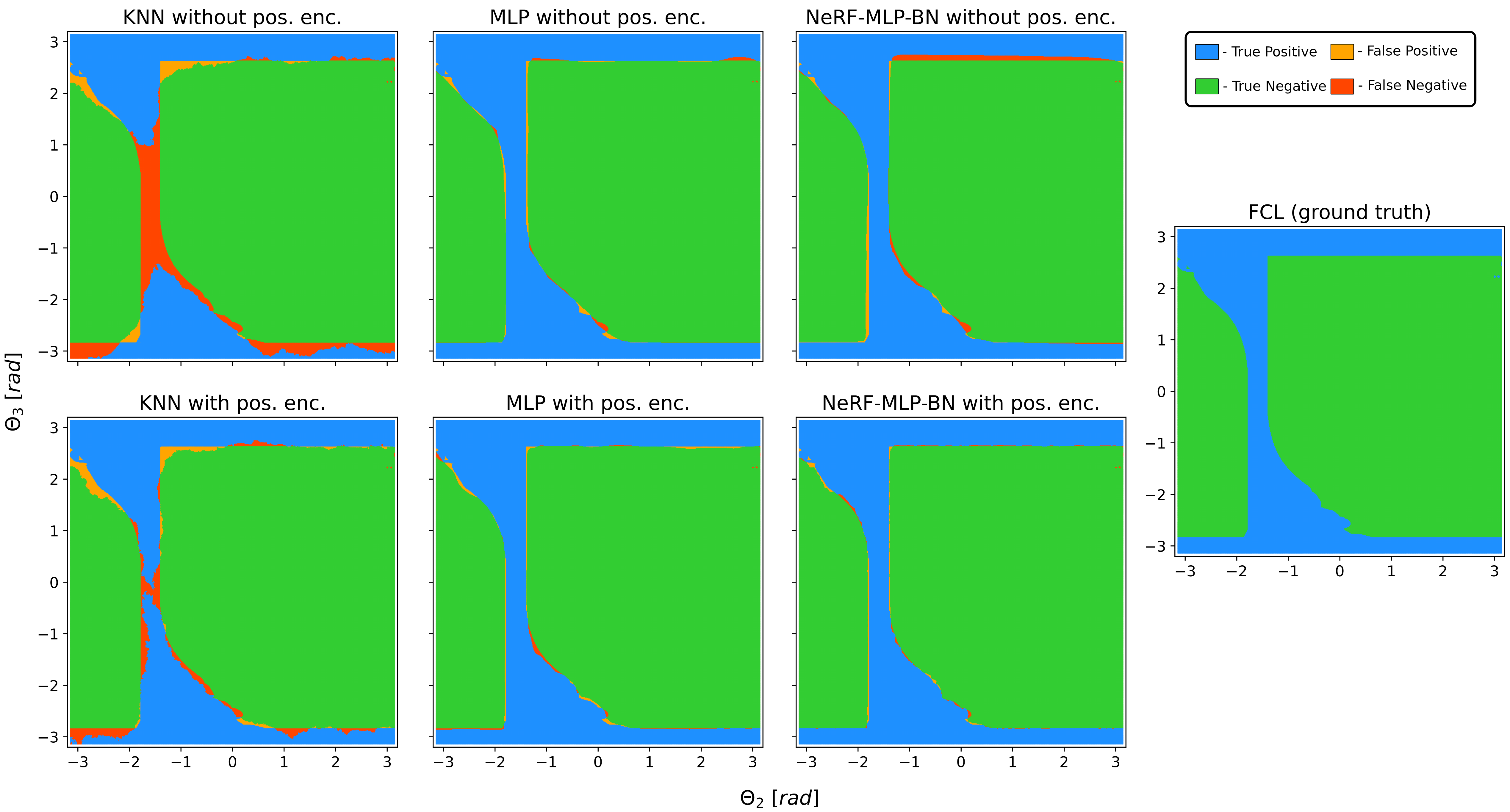}
    \caption{Visualization of the predicted collision state in the robot workspace for the second and third joints obtained using K-Nearest Neighbors (KNN), Multilayer Perceptron (MLP3), and ${\rm NeRF}_{\rm MLP}$ model with BN. The top row shows results obtained after training with the raw input values, while the bottom row shows results for the encoded input -- both compared to the FCL (ground truth).}
    \label{fig:configuration_space}
\end{figure}

In addition to evaluating accuracy, we illustrate the impact of positional encoding by comparing collision state predictions within the robot’s configuration space. Figure~\ref{fig:configuration_space} presents the self-collision states detected by various methods for configurations with varying joint angles $\theta_2$ and $\theta_3$, while the remaining joint angles are set to zero. The ground truth, used for comparison, is based on results from the standard collision detection method — the Flexible Collision Library (FCL) used in ROS.

For the K-Nearest Neighbors classifier, positional encoding reduces errors in detecting collision states near the boundaries of the collision-free space. The two collision-free areas were almost completely separated as required, and the overall shape more closely resembles the ground truth. The improvement is less pronounced for the neural network-based classifiers (MLP and NeRF). However, positional encoding still enhances the model’s ability to capture higher-frequency features within the configuration space. For example, the ground truth collision-free space has an intricate structure in regions around $\theta_2{=}-2.8, \theta_3{=}2.3$ and $\theta_2{=}0.0, \theta_3{=}-2.5$.
Neural networks trained with positional encoding may generate sharper class boundaries and thus produce more accurate predictions for complex shapes like those visible in the mentioned regions.

\subsection{Training performance and inference time}
\begin{figure}[t]
    \centering
    \includegraphics[width=0.499\textwidth]{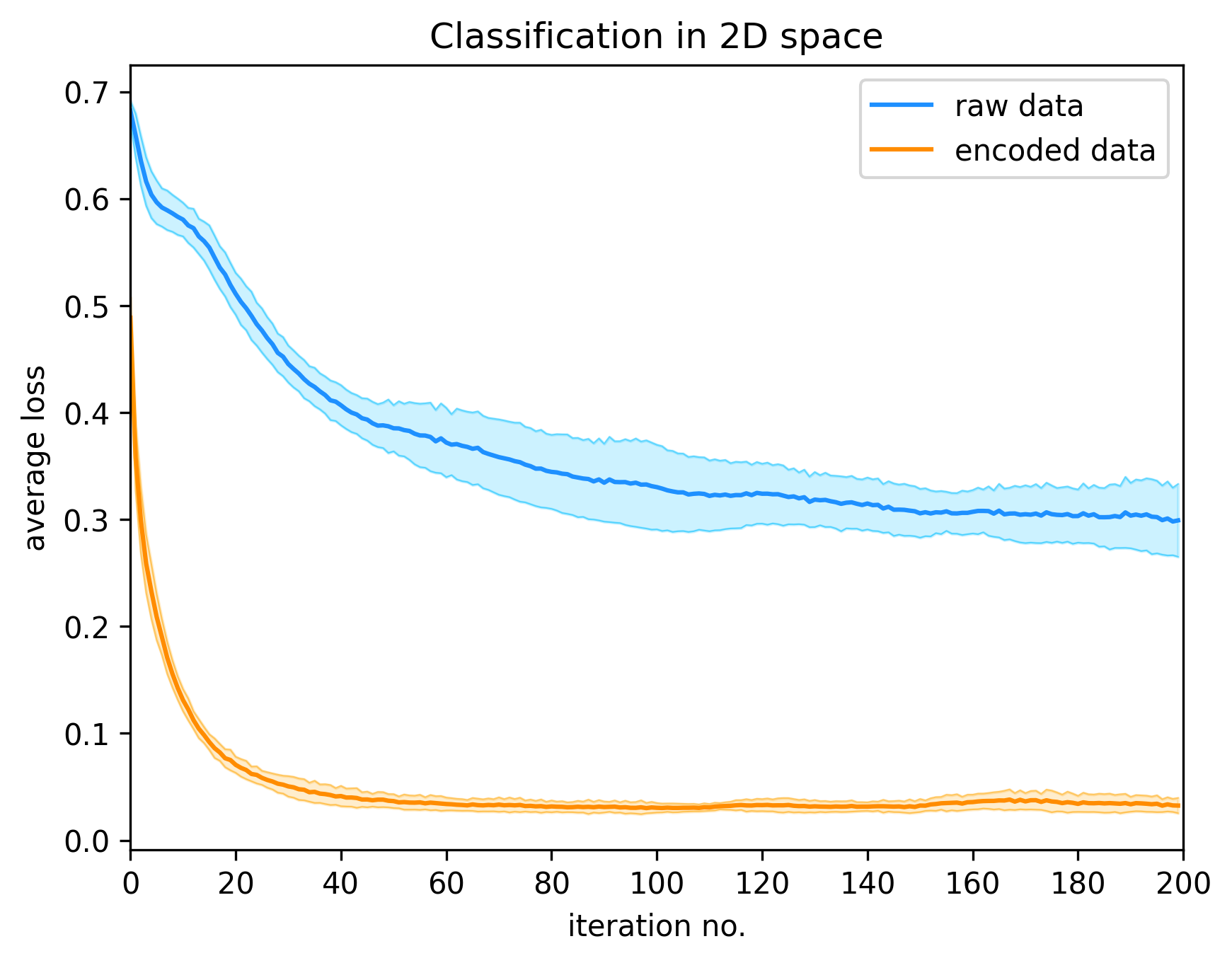}%
    \includegraphics[width=0.499\textwidth]{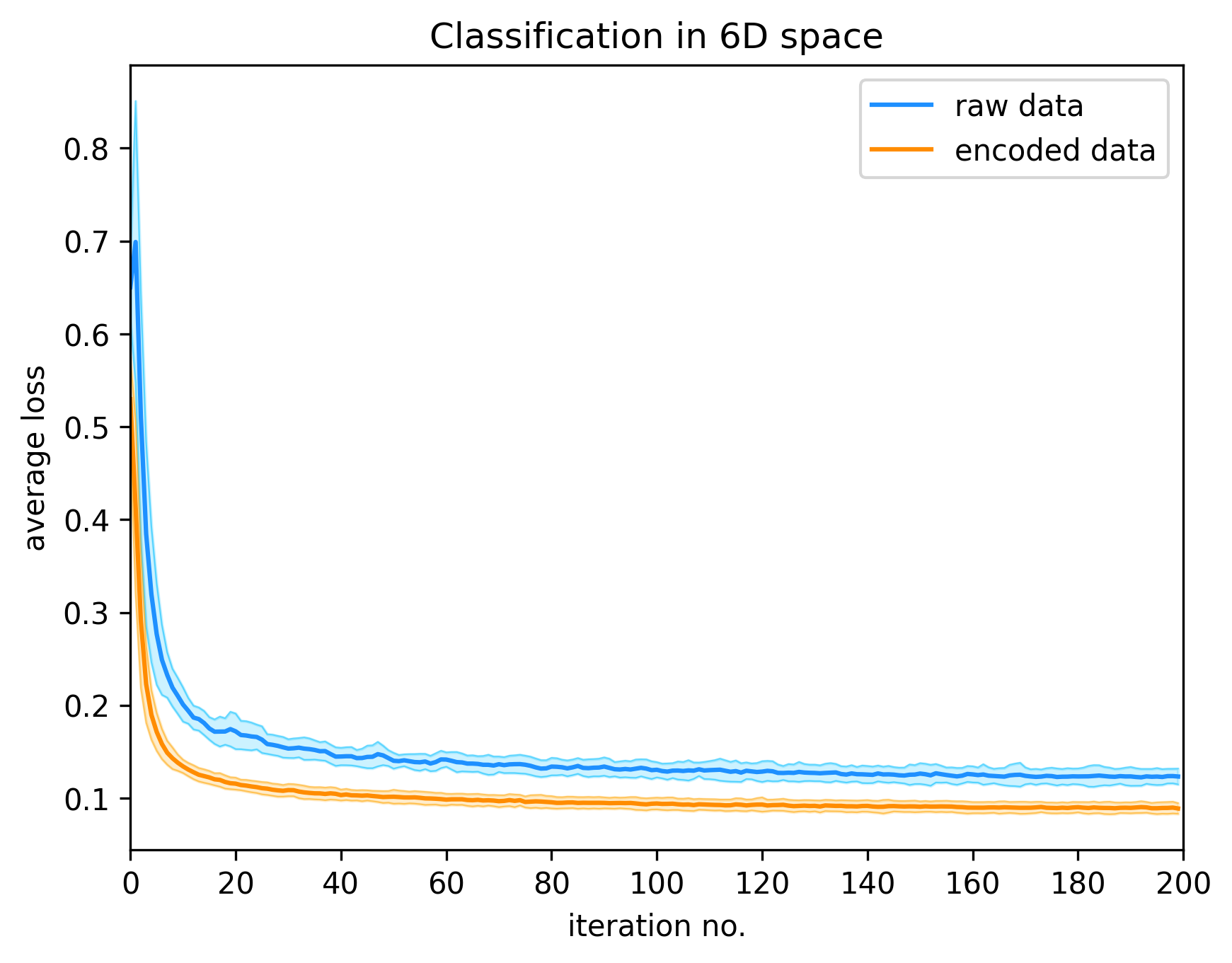}
    \caption{Average loss curves (based on 10 train trials) for training classifier in 2D space (x, y coordinates) and 6D space (robot joint values) using input with and without encoding.}
    \label{fig:loss_curves}
\end{figure}

Another interesting feature of positional encoding is its influence on the training process. To investigate this, we compared the learning curves when training the model using raw or encoded input data, as shown in Fig.~\ref{fig:loss_curves}. The curves represent the average cross-entropy loss. We can observe a relatively slow training process for two-dimensional raw input data. Applying the positional encoding (with L=5) results in much faster convergence and higher repeatability (standard deviation reduction). We conducted the same experiment for robot joint data (6D) -- raw and encoded with L=3, using MLP3 neural network architecture. The figure shows less improvement than for 2D space, but we can still observe significant training stabilization. The gap between loss curves for raw and encoded data emphasizes improvement in final model accuracy.

\begin{figure*}[t]
    \centering
    \includegraphics[width=0.255\textwidth]{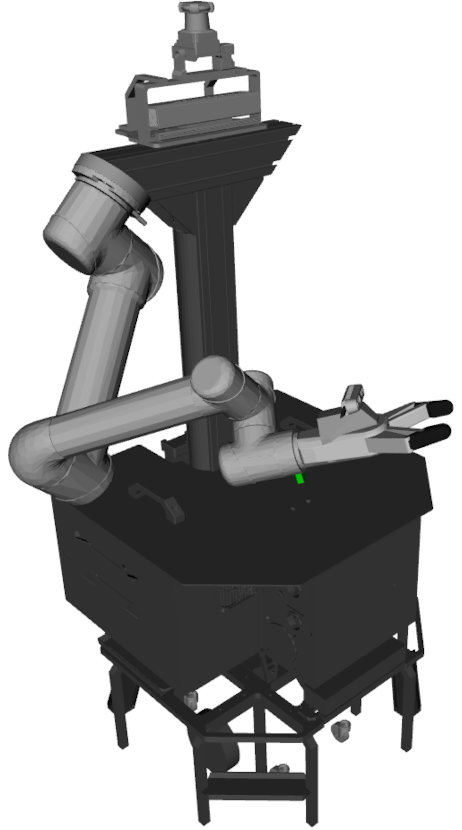}%
    \includegraphics[width=0.245\textwidth]{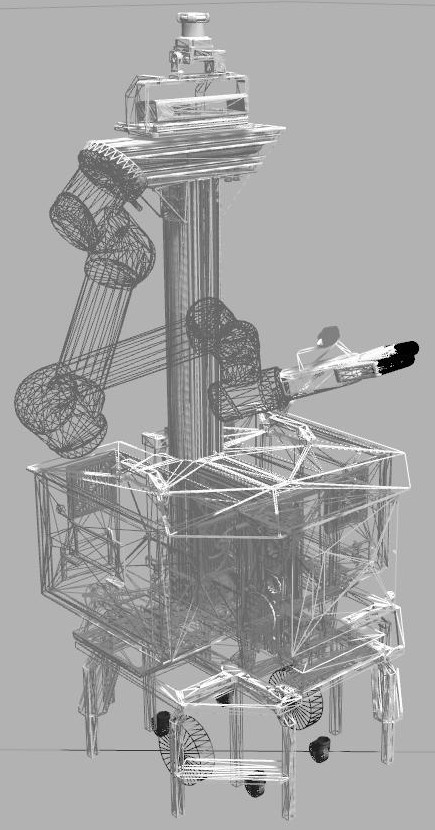}%
    \includegraphics[width=0.255\textwidth]{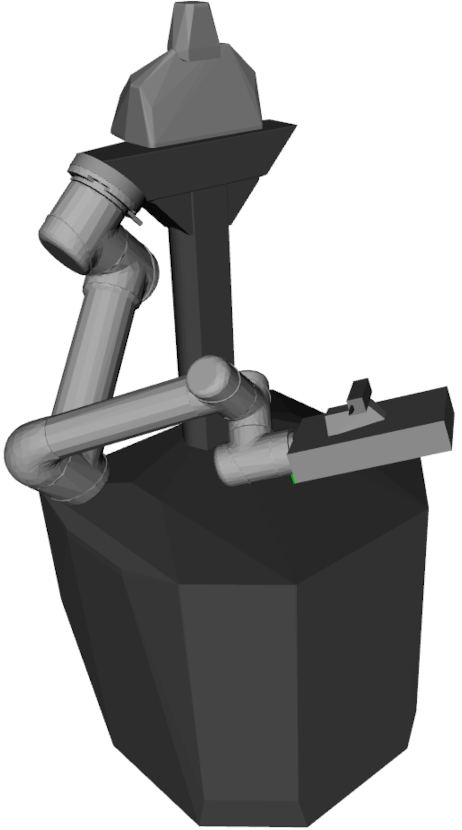}%
    \includegraphics[width=0.245\textwidth]{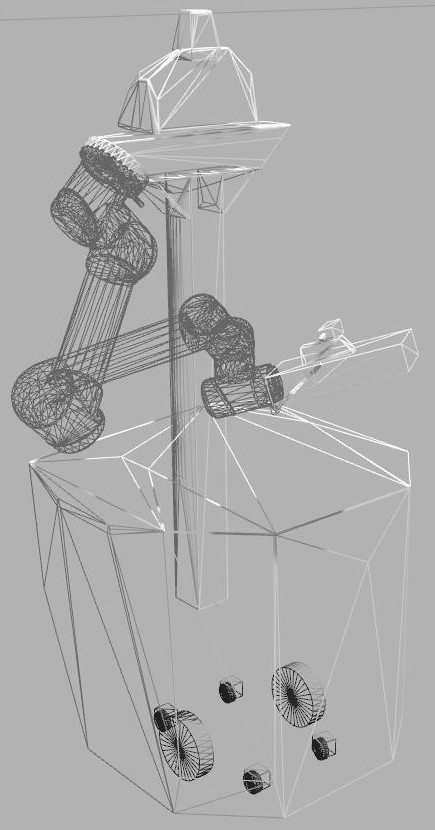}
    \put(-366,1){a} \put(-180,1){b}
    \caption{Robot models used for collision checking in FCL and their wireframe views: precise robot model (ground truth) (a) and simplified robot model (b).}
    \label{fig:fcl_models}
\end{figure*}

In the introduction section, we mention that neural networks outperform geometric-based methods in terms of self-collision checking time. To prove this, we checked the average inference time\footnote{Time measurements were performed using ONNX Runtime on standard PC with Intel\textsuperscript{\scalebox{.7}{\textregistered}} Core\textsuperscript{\scalebox{.7}{TM}} i7-9750H CPU. } for the test dataset using all developed neural networks, the best conventional classifier, and the standard geometric-based methods. 
The ground truth data relies on the output of FCL operating on a precise robot model. The simplified robot model was prepared for better comparison, consisting of selected mesh convex hulls or basic shapes like boxes and cylinders. Both models are illustrated in Fig.~\ref{fig:fcl_models}. The figure shows also the wireframe view of the models highlighting the reduced number of triangles in the simplified version (from 91k to 7k).

\begin{figure*}[t!]
    \centering
    \includegraphics[width=0.7\textwidth]{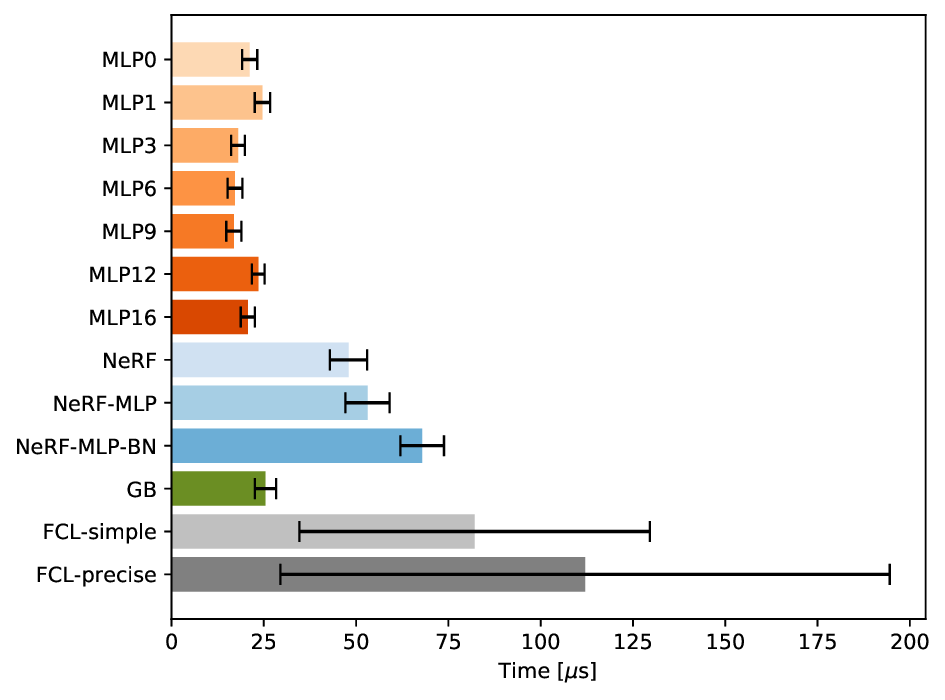}
    \caption{Mean inference time for various classifier architectures obtained using all tested input vector lengths compared to FCL operating on a precise robot model (ground truth) and FCL operating on a simplified robot model.}
    \label{fig:times}
\end{figure*}

The self-collision checking time values are presented in Fig.~\ref{fig:times}. For data-driven methods, the mean time is calculated for the whole range of input vector lengths -- from L=0 (no positional encoding) to L=20. 
All machine learning models reach shorter mean inference times than collision checking using FCL. 
FCL operating on a simple robot model results in 27\% shorter time than for a precise FCL version, but causes false positives to occur -- the mean accuracy is 0.947. 
The fastest model is MLP9. It checks self-collision for a single robot state 6.6 times faster than FCL (for precise model). The inference time for neural networks depends on their size presented in Tab.~\ref{tab:architectures_mlp}. Models from the NeRF family have about ten hidden layers (vs. 3 for MLPs), so their inference time is significantly higher. The impact of the positional encoding augmentation level on the time is relatively small -- it fits within the error bars (standard deviation). 
The neural network inference time can be further reduced using batch processing. The experiments show that feeding the MLP with a batch of 100 samples results in about 38 times faster inference than with sequential sample processing. Such an approach may result in a significant motion planning time reduction.
Another benefit of machine learning collision checkers is the independence of inference time from the robot configuration. For the geometric-based method, checking the collision in demanding robot configurations (with many mesh contacts) takes many times longer than for those without colliding parts. As a result, we observe a high standard deviation of FCL inference time -- it is less predictable than for ML models.

The results show that GB is nearly as fast as the slowest MLP while being more accurate than the best one. When selecting between GB and neural networks for self-collision checking, there is a trade-off between speed and accuracy. However, neural networks offer an additional advantage: their differentiability enables gradient-based planners to optimize trajectories and avoid self-collisions. This capability is not feasible when using GB for self-collision checking.

\section{Conclusions}

In this article, we applied machine learning methods to the robot self-collision detection task. Such a data-driven approach outperforms the classic geometric method in the state validation time and time-to-state independence. These methods achieve high efficiency, which can be boosted using an input encoding. We show the improvement of using the positional encoding by 1.317\% for MLP, 2.274\% for the NeRF-based architecture, and 10.2\% for the 2D synthetic problem. Moreover, the proposed neural network-based self-collision detectors are differentiable in contrast to classical methods like mesh-based collision detection or Gradient Boosting. It means that the obtained model can be used to check constraints and also to quickly determine the direction that allows for avoiding collision.

The article demonstrates how classification accuracy changes with different values of the positional encoding parameter $L$. 
We evaluated multiple neural network architectures and performed similar analyses on traditional machine learning models. Our findings show that augmenting the feature vector with positional encoding consistently improves classification performance by about 1\%. This enhancement is evident across various neural network architectures, including MLP and NeRF-based models, as well as other machine learning techniques. An important factor is a proper encoding level ($L$) determination to obtain accuracy improvement while maintaining a compact input vector. In the case of multidimensional robot configuration space (6D), the improvement level is not as distinct as for 2D space, but it is still significant for most models. We can observe higher accuracy and also the ability to model higher-frequency features.

In the future, we are going to develop techniques that adapt the frequency of the sine and cosine functions during training. We believe that an additional degree of freedom during training would allow for better adaptation of the positional encoding to the properties of the input data and further improve the classification results. We are also going to investigate 3D Gaussian Splatting models for efficient collision checking.

\section*{Acknowledgment}
The work was supported by the National Science Centre, Poland, under research project no  UMO-2023/51/B/ST6/01646.

\bibliographystyle{plain}
\bibliography{IEEEexample}

\end{document}